arXiv:1709.07093v1 [cs.LG] 20 Sep 2017

# Near Optimal Sketching of Low-Rank Tensor Regression

Jarvis Haupt, Xingguo Li and David P. Woodruff*

**Abstract**

We study the least squares regression problem

$$\min_{\Theta \in \mathcal{S}_{\odot D,R}} \|A\Theta - b\|_2,$$

where $\mathcal{S}_{\odot D,R}$ is the set of $\Theta$ for which $\Theta = \sum_{r=1}^{R} \theta_1^{(r)} \circ \cdots \circ \theta_D^{(r)}$ for vectors $\theta_d^{(r)} \in \mathbb{R}^{p_d}$ for all $r \in [R]$ and $d \in [D]$, and $\circ$ denotes the outer product of vectors. That is, $\Theta$ is a low-dimensional, low-rank tensor. This is motivated by the fact that the number of parameters in $\Theta$ is only $R \cdot \sum_{d=1}^{D} p_d$, which is significantly smaller than the $\prod_{d=1}^{D} p_d$ number of parameters in ordinary least squares regression. We consider the above CP decomposition model of tensors $\Theta$, as well as the Tucker decomposition. For both models we show how to apply data dimensionality reduction techniques based on *sparse* random projections $\Phi \in \mathbb{R}^{m \times n}$, with $m \ll n$, to reduce the problem to a much smaller problem $\min_{\Theta} \|\Phi A\Theta - \Phi b\|_2$, for which if $\Theta'$ is a near-optimum to the smaller problem, then it is also a near optimum to the original problem. We obtain significantly smaller dimension and sparsity in $\Phi$ than is possible for ordinary least squares regression, and we also provide a number of numerical simulations supporting our theory.

## 1 Introduction

For a sequence of $D$-way design tensors $A_i \in \mathbb{R}^{p_1 \times \cdots \times p_D}$, $i \in [n] = \{1, \ldots, n\}$, we observe noisy linear measurements of an unknown $D$-way tensor $\Theta \in \mathbb{R}^{p_1 \times \cdots \times p_D}$, given by

$$b_i = \langle A_i, \Theta \rangle + z_i, \text{ for all } i \in [n], \tag{1}$$

where $\{z_i\}_{i=1}^n$ corresponds to the noise in each observation, and $\langle A_i, \Theta \rangle = \text{vec}(A_i)^\top \text{vec}(\Theta)$, with $\text{vec}(X)$ denoting the vectorization of a tensor $X$. Given the design tensors $\{A_i\}_{i=1}^n$ and noisy observations $\{b_i\}_{i=1}^n$, a natural approach for estimating the parameter $\Theta$ is to use the *Ordinary Least Square* (OLS) estimation for tensor regression, i.e., to solve

$$\min_{\Theta \in \mathbb{R}^{p_1 \times \cdots \times p_D}} \sum_{i=1}^{n} \left(b_i - \langle A_i, \Theta \rangle\right)^2. \tag{2}$$

*The authors are listed in the alphabetical order. Correspondence to: Xingguo Li <lixx1661@umn.edu>. The paper is Accepted at NIPS 2017. The authors acknowledge support from University of Minnesota Startup Funding. Xingguo Li and Jarvis Haupt are affiliated with Department of Electrical and Computer Engineering at University of Minnesota, Minneapolis, MN, 55455, USA; Xingguo Li is also affiliated with School of Industrial and Systems Engineering at Georgia Institute of Technology; David P. Woodruff is affiliated with School of Computer Science, Carnegie Mellon University, Pittsburgh, PA, 95120, USA.

Tensor regression has been widely studied in the literature. Applications include computer vision (Park and Savvides, 2007; Guo et al., 2012; Zhao et al., 2013), data mining (De Lathauwer et al., 2000), multi-model ensembles (Yu et al., 2015), neuroimaging analysis (Zhou et al., 2013; Li et al., 2013b), multitask learning (Romera-Paredes et al., 2013; Yang and Hospedales, 2016), and multivariate spatial-temporal data analysis (Bahadori et al., 2014; Hoff, 2015). In these applications, modeling the unknown parameters as a tensor is what is needed, as it allows for learning data that has multi-directional relations, such as in climate prediction Yu and Liu (2016), inherent structure learning with multi-dimensional indices Romera-Paredes et al. (2013), and hand movement trajectory decoding Zhao et al. (2013).

Due to the high dimensionality of tensor data, structured learning based on low-rank tensor decompositions, such as CANDECOMP/PARAFAC (CP) decomposition and Tucker decomposition models (Kolda and Bader, 2009; Sidiropoulos et al., 2016) have been proposed in order to obtain tractable tensor regression problems. As discussed more below, requiring the unknown tensor to be low-rank significantly reduces the number of unknown parameters. As natural convex formulations based on the nuclear norm are known to be computationally expensive (Gandy et al., 2011; Tomioka and Suzuki, 2013), nonconvex heuristics for low-rank tensor recovery are often used in practice (Romera-Paredes et al., 2013; Bahadori et al., 2014; Yu et al., 2015).

We consider low-rank tensor regression problems based on the CP decomposition and Tucker decomposition models. For simplicity, we first focus on the CP model, and later extend our analysis to the Tucker model. Suppose that $\Theta$ admits a rank-$R$ CP decomposition, that is,

$$\Theta = \sum_{r=1}^{R} \theta_1^{(r)} \circ \cdots \circ \theta_D^{(r)}, \tag{3}$$

where $\theta_d^{(r)} \in \mathbb{R}^{p_d}$ for all $r \in [R]$ and $\circ$ is the outer product of vectors. For convenience, we denote the set of factors for low-rank tensors by

$$\mathcal{S}_{D,R} = \Big\{ [[\Theta_1, \ldots, \Theta_D]] \mid \Theta_d = [\theta_d^{(1)}, \ldots, \theta_d^{(R)}] \in \mathbb{R}^{p_d \times R}, \text{ for all } d \in [D] \Big\}.$$

Then we can rewrite model (1) in a compact form

$$b = A(\Theta_D \odot \cdots \odot \Theta_1) 1_R + z, \tag{4}$$

where $b, z \in \mathbb{R}^n$, $A = [\mathrm{vec}(A_1), \cdots, \mathrm{vec}(A_n)]^\top \in \mathbb{R}^{n \times \prod_{d=1}^{D} p_d}$, $1_R = [1, \ldots, 1] \in \mathbb{R}^R$ is a vector of all 1s, $\otimes$ is the Kronecker product, and $\odot$ is the Khatri-Rao product[1]. In addition, the OLS estimation for tensor regression (2) can be rewritten as the following nonconvex problem in terms of low-rank tensor parameters $[[\Theta_1, \ldots, \Theta_D]]$,

$$\min_{\vartheta \in \mathcal{S}_{\odot D,R}} \|A\vartheta - b\|_2^2, \tag{5}$$

where $\mathcal{S}_{\odot D,R} = \Big\{ (\Theta_D \odot \cdots \odot \Theta_1) 1_R \in \mathbb{R}^{\prod_D^{d=1} p_d} \mid [[\Theta_1, \ldots, \Theta_D]] \in \mathcal{S}_{D,R} \Big\}$.

[1] These are defined below in the section of notation.

The number of parameters for a general tensor $\Theta \in \mathbb{R}^{p_1 \times \cdots \times p_D}$ is $\prod_{d=1}^{D} p_d$, which may be prohibitive even for small values of $\{p_d\}_{d=1}^{D}$. The benefit of the low-rank model (3) is that it dramatically reduces the degrees of freedom of the unknown tensor from $\prod_{d=1}^{D} p_d$ to $R\sum_{d=1}^{D} p_d$, where we are typically interested in the case when $R \ll p_d$ for each $d \in [D]$. For example, a typical MRI image has size $256^3 \approx 1.7\times 10^7$, while using the low-rank model with $R = 5$, we reduce the number of unknown parameters to $256\times 3\times 5 \approx 4\times 10^3 \ll 10^7$. This significantly increases the applicability of the tensor regression model in practice.

Nevertheless, solving the tensor regression problem (5) is still expensive in terms of both computation and memory requirements, for typical settings, when $n \gg R \cdot \sum_{d=1}^{D} p_d$, or even $n \gg \prod_{d=1}^{D} p_d$. In particular, the per iteration complexity is at least linear in $n$ for popular algorithms such as block alternating minimization and block gradient descent (Tseng, 2001; Tseng and Yun, 2009). In addition, in order to store $A$, it takes $n \cdot \prod_{d=1}^{D} p_d$ words of memory. Both of these aspects are undesirable when $n$ is large. This motivates us to consider data dimensionality reduction techniques, also called *sketching*, for the tensor regression problem.

Instead of solving (5), we consider the *Sketched Ordinary Least Square* (SOLS) estimation problem, defined as

$$\min_{\vartheta \in \mathcal{S}_{\odot D,R}} \|\Phi A\vartheta - \Phi b\|_2^2, \tag{6}$$

where $\Phi \in \mathbb{R}^{m\times n}$ is a random matrix specified below. Importantly, $\Phi$ will satisfy two properties discussed below, namely (1) $m \ll n$ so that we significantly reduce the size of the problem, and (2) $\Phi$ will be very sparse so that it can be applied very quickly.

Naïvely applying existing analyses of sketching techniques for least squares regression requires $m = \Omega\left(\prod_{d=1}^{D} p_d\right)$ (for a survey, see, e.g., (Woodruff, 2014)), which is prohibitive. Here, we use a sparse Johnson-Lindenstrauss transformation (SJLT) as our sketching matrix, with constant column sparsity and dimension $m = \Omega\left(R \cdot \sum_{d=1}^{D} p_d\right)$, up to logarithmic factors. We show that with high probability, simultaneously for every $\vartheta \in \mathcal{S}_{\odot,D,R}$, we have $\|\Phi A\vartheta - \Phi b\|_2^2 = (1 \pm \epsilon)\|A\vartheta - b\|_2^2$, which implies that any solution to (6) has the same cost as in (5) up to a $(1+\epsilon)$-factor. In particular, by solving (6) we obtain a $(1+\epsilon)$-approximation to (5). Our result is the first non-trivial dimensionality reduction for this problem, i.e., dimensionality reduction better than $\left(\prod_{d=1}^{D} p_d\right)$, which is trivial by ignoring the low rank structure of the tensor, and which achieves a relative error $(1+\epsilon)$-approximation.

Our analysis is based on a careful characterization of Talagrand's functional for the parameter space of low-rank tensors. Our sketching dimension $m$ almost meets the intrinsic dimension of low-rank tensors, and is thus nearly optimal. We further provide numerical evaluations on both synthetic and real data to demonstrate the empirical performance of sketching based estimation.

**Notation**. For scalars $x, y \in \mathbb{R}$, we denote $x = (1 \pm \varepsilon)y$ if $x \in [(1-\varepsilon)y, (1+\varepsilon)y]$, $x \lesssim (\gtrsim) y$ if $x \le (\ge) cy$ for some universal constant $c > 0$, and $x \simeq y$ if both $x \lesssim y$ and $x \gtrsim y$ hold. We also use standard asymptotic notation $\mathcal{O}(\cdot)$ and $\Omega(\cdot)$. Given a positive integer $n$, let $[n] = \{1, \ldots, n\}$. Given a vector $v \in \mathbb{R}^p$, we denote $\|v\|_1 = \sum_{i=1}^{p} |v_i|$, $\|v\|_2^2 = \sum_{i=1}^{p} v_i^2$, and $\|v\|_\infty = \max_{i\in[p]} |v_i|$. Given $d$ vectors $v_1 \in \mathbb{R}^{p_1}, \ldots, v_d \in \mathbb{R}^{p_d}$, we denote $v_1 \circ \cdots \circ v_d \in \mathbb{R}^{p_1\times\cdots\times p_d}$ as a tensor formed by the outer product of vectors. Given a matrix $A \in \mathbb{R}^{m\times n}$, we denote its spectral norm by $\|A\|_2$, we let $\operatorname{span}(A) \subseteq \mathbb{R}^m$ be the subspace spanned by the columns of $A$, we let $\sigma_{\max}(A)$ and $\sigma_{\min}(A)$ be the largest and smallest singular values of $A$, respectively, and $\kappa(A) = \sigma_{\max}(A)/\sigma_{\min}(A)$ be the condition number. We use $\text{nnz}(A)$ to denote the number of nonzero entries of $A$. We use $\mathcal{P}_A$ as the projection operator onto $\text{span}(A)$. Given two matrices $A = [a_1, \ldots, a_n] \in \mathbb{R}^{m\times n}$ and $B = [b_1, \ldots, b_q] \in \mathbb{R}^{p\times q}$, $A\otimes B = [a_1\otimes B, \ldots, a_n\otimes B] \in \mathbb{R}^{mp\times nq}$ denotes the Kronecker product, and $A \odot B = [a_1 \otimes b_1, \ldots, a_n \otimes b_n] \in \mathbb{R}^{mp\times n}$ denotes the Khatri-Rao product with $n = q$. We let $\mathcal{B}_n \subset \mathbb{R}^n$ be the unit sphere of $\mathbb{R}^n$, i.e., $\mathcal{B}_n = \{x \in \mathbb{R}^n \mid \|x\|_2 = 1\}$. We also let $\mathbb{P}(\cdot)$ be the probability of an event and $\mathbb{E}(\cdot)$ the expectation of a random variable. Without further specification, we let $\prod p_d = \prod_{d=1}^D p_d$ and $\sum p_d = \sum_{d=1}^D p_d$.

## 2 Background

We start with a few important definitions.

**Definition 1** (Oblivious Subspace Embedding)**.** Suppose $\Pi$ is a distribution on $m \times n$ matrices $\Phi$, where $m$ is a function of parameters $n, d$, and $\varepsilon$. Further suppose that with probability at least $1 - \delta$, for any fixed $n \times d$ matrix $A$, a matrix $\Phi$ drawn from $\Pi$ has the property that $\Phi$ is a $(1 \pm \varepsilon)$ subspace embedding for $A$, i.e., $\|\Phi Ax\|_2^2 = (1 \pm \varepsilon)\|Ax\|_2^2$ for any $x \in \mathcal{X} \subseteq \mathbb{R}^d$. Then we call $\Pi$ an $(\varepsilon, \delta)$ *oblivious subspace embedding* (OSE) of $\mathcal{X}$.

An OSE $\Phi$ preserves the norm of vectors in a certain set $\mathcal{X}$ after linear transformation by $A$. This is widely studied as a key property for sketching based analyses (see (Woodruff, 2014) and the references therein). We want to show an analogous property when $\mathcal{X}$ is parameterized by a low-rank tensor model.

**Definition 2** (Leverage Scores)**.** Given $A \in \mathbb{R}^{n\times d}$, let $Z \in \mathbb{R}^{n\times d}$ have orthonormal columns that span the column space of $A$. Then $\ell_i^2(A) = \|e_i^\top Z\|_2^2$ is the $i$-th *leverage score* of $A$.

Leverage scores play an important role in randomized matrix algorithms (Mahoney and Drineas, 2009; Mahoney, 2011; Drineas et al., 2012). Calculating the leverage scores naïvely by orthogonalizing $A$ requires $\mathcal{O}(nd^2)$ time. It is shown in Clarkson and Woodruff (2013) that the leverage scores of $A$ can be approximated individually up to a constant multiplicative factor in $\mathcal{O}(\text{nnz}(A)\log n + \text{poly}(d))$ time using sparse subspace embeddings.

**Definition 3** (Sparse Johnson-Lindenstrauss Transforms)**.** Let $\sigma_{ij}$ be independent Rademacher random variables, i.e., $\mathbb{P}(\sigma_{ij} = 1) = \mathbb{P}(\sigma_{ij} = -1) = 1/2$, and let $\delta_{ij} : \Omega_\delta \to \{0, 1\}$ be random variables, independent of the $\sigma_{ij}$, with the following properties:

- $\delta_{ij}$ are negatively correlated for fixed $j$, i.e., for all $1 \le i_1 < \ldots < i_k \le m$,

$$\mathbb{E}\left(\prod_{t=1}^k \delta_{i_t,j}\right) \le \prod_{t=1}^k \mathbb{E}\left(\delta_{i_t,j}\right) = \left(\frac{s}{m}\right)^k;$$

- There are $s = \sum_{i=1}^m \delta_{ij}$ nonzero $\delta_{ij}$ for a fixed $j$;
- The vectors $(\delta_{ij})_{i=1}^m$ are independent across $j \in [n]$.

Then $\Phi \in \mathbb{R}^{m\times n}$ is a *sparse Johnson-Lindenstrauss transform* (SJLT) matrix if $\Phi_{ij} = \frac{1}{\sqrt{s}}\sigma_{ij}\delta_{ij}$.

The SJLT has several benefits (Dasgupta et al., 2010; Kane and Nelson, 2014; Woodruff, 2014). First, the computation of $\Phi x$ takes only $\mathcal{O}(\mathrm{nnz}(x))$ time when $s$ is a constant. Second, storing $\Phi$ takes only $sn$ memory instead of $mn$, which is significant when $s \ll m$. This can often further be reduced by drawing the entries of $\Phi$ from a limited independent family of random variables. We will use an SJLT as the sketching matrix in our analysis and our goal will be to show sufficient conditions on $m$ and $s$ such that the analogue of the OSE property holds for low-rank tensor regression.

**Definition 4** (Talagrand's Functional)**.** Given a (semi-)metric $\rho$ on $\mathbb{R}^n$ and a bounded set $\mathcal{S} \subset \mathbb{R}^n$, *Talagrand's $\gamma_2$-functional* is

$$\gamma_2(S,\rho) = \inf_{\{\mathcal{S}_r\}_{r=0}^{\infty}} \sup_{x\in S} \sum_{r=0}^{\infty} 2^{r/2}\cdot \rho(x,\mathcal{S}_r), \tag{7}$$

where $\rho(x,\mathcal{S}_r)$ is a distance from $x$ to $\mathcal{S}_r$ and the infimum is taken over all collections $\{\mathcal{S}_r\}_{r=0}^{\infty}$ such that $\mathcal{S}_0 \subset \mathcal{S}_1 \subset \ldots \subset \mathcal{S}$ with $|\mathcal{S}_0|= 1$ and $|\mathcal{S}_r|\leq 2^{2^r}$.

A closely related notion of $\gamma_2$-functional is the *Gausssian mean width*,

$$\mathcal{G}(\mathcal{S}) = \mathbb{E}_g \sup_{x\in\mathcal{S}} \langle g,x\rangle,$$

where $g \sim \mathcal{N}_n(0,I_n)$. For any bounded $\mathcal{S} \subset \mathbb{R}^n$, $\mathcal{G}(\mathcal{S})$ and $\gamma_2(\mathcal{S},\|\cdot\|_2)$ differ multiplicatively by at most a universal constant in Euclidean space. Both of these quantities are widely used (Talagrand, 2006). Finding a tight upper bound on the $\gamma_2$-functional for the parameter space of low-rank tensors is a key part of our analysis.

**Definition 5** (Finsler Metric)**.** Let $E,E' \subset \mathbb{R}^n$ be $p$-dimensional subspaces. The *Finsler metric* of $E$ and $E'$ is

$$\rho_{\mathrm{Fin}}(E,E') = \|\mathcal{P}_E - \mathcal{P}_{E'}\|_2,$$

where $\mathcal{P}_E$ is the projection onto the subspace $E$.

The Finsler metric is the semi-metric used in the $\gamma_2$-functional in our analysis. Note that $\rho_{\mathrm{Fin}}(E,E') \leq 1$ always holds for any $E$ and $E'$ (Shen, 2001). See further discussion in Section 3.

# 3 Dimensionality Reduction for CP Decomposition

For convenience, we introduce the following notation. Given a $D$-way tensor $\Theta = \sum_{r=1}^{R} \theta_1^{(r)} \circ \cdots \circ \theta_D^{(r)} \in \mathbb{R}^{p_1\times\cdots\times p_D}$, where $\theta_d^{(r)} \in \mathbb{R}^{p_d}$ for all $d \in [D]$ and $r \in [R]$, we consider fixing all but $\theta_1^{(r)}$ for $r \in [R]$, and denoting

$$A^{\left\{\theta_{\backslash 1}^{(r)}\right\}} = \left[A^{\theta_{\backslash 1}^{(1)}},\ldots,A^{\theta_{\backslash 1}^{(R)}}\right] \in \mathbb{R}^{n\times Rp_1},$$

where

$$A^{\theta^{(i)}_{\backslash 1}} = \sum_{j_D=1}^{p_D} \cdots \sum_{j_2=1}^{p_2} A^{(j_D,\ldots,j_2)} \cdot \theta^{(i)}_{D,j_D} \cdots \theta^{(i)}_{2,j_2}$$

$$A = \left[A^{(1,\ldots,1)}, A^{(1,\ldots,2)}, \ldots, A^{(p_D,\ldots,p_2)}\right] \in \mathbb{R}^{n\times \prod p_d}$$

$\theta^{(i)}_{d,j_d}$ is the $j_d$-th entry of $\theta^{(i)}_d$, and $A^{(j_D,\ldots,j_2)} \in \mathbb{R}^{n\times p_1}$ for all $j_d \in [p_d]$, $d \in [D]\backslash\{1\}$. The above parameterization allows us to view tensor regression as preserving the norms of vectors in an infinite union of subspaces, described in more detail below.

We provide sufficient conditions for the SJLT matrix $\Phi \in \mathbb{R}^{m\times n}$ to preserve the cost of all solutions for tensor regression, i.e., bounds on the sketching dimension $m$ and the per-column sparsity $s$ for which

$$\mathop{\mathbb{E}}_{\Phi} \sup_{x\in\mathcal{T}} \left| \|\Phi x\|_2^2 - 1 \right| < \varepsilon \tag{8}$$

where $\varepsilon$ is a given precision, $\mathcal{T} = \bigcup_{E\in\mathcal{V}} \{x \in E \mid \|x\|_2 = 1\}$, and

$$\mathcal{V} = \bigcup_{\theta^{(r)}_d \in \mathbb{R}^{p_d}, \forall r\in[R],\ d\in[D]\backslash\{1\}} \operatorname{span}\left[A^{\left\{\theta^{(r)}_{\backslash 1}\right\}}, A^{\left\{\phi^{(r)}_{\backslash 1}\right\}}\right].$$

Note that by linearity, it suffices to consider $x$ with $\|x\|_2 = 1$ in the above, which explains the form of (8). Also note that (8) implies for all $\vartheta \in \mathcal{S}_{\odot D,R}$, that

$$\|\Phi A\vartheta - \Phi b\|_2^2 = (1 \pm \varepsilon)\|A\vartheta - b\|_2^2, \tag{9}$$

which allows us to minimize the much smaller sketched problem to obtain parameters $\vartheta$ which, when plugged into the original objective function, provide a multiplicative $(1+\epsilon)$-approximation.

We need the following theorem for embedding an infinite union of subspaces. All proofs can be found in the supplementary material.

**Theorem 1.** Let $\mathcal{T} \subset \mathcal{B}_n$ and $\Phi \in \mathbb{R}^{m\times n}$ be an SJLT matrix with column sparsity $s$, and

$$p_{\mathcal{V}} = \sup_{\substack{\theta^{(r)}_d \in \mathbb{R}^{p_d}, \forall r\in[R],\\ d\in[D]\backslash\{1\}}} \dim\left(\operatorname{span}\left[A^{\left\{\theta^{(r)}_{\backslash 1}\right\}}, A^{\left\{\phi^{(r)}_{\backslash 1}\right\}}\right]\right).$$

Then (8) holds if $m$ and $s$ satisfy

$$m \gtrsim \varepsilon^{-2}(\log^4 m)(\log^5 n)\left(\gamma_2^2(\mathcal{V}, \rho_{\mathrm{Fin}}) + p_{\mathcal{V}} + \log \mathcal{N}(\mathcal{V}, \rho_{\mathrm{Fin}}, \varepsilon_0)\right), \tag{10}$$

$$s \gtrsim \varepsilon^{-2}(\log^6 m)(\log^5 n)\left(\left[\int_0^{\varepsilon_0} (\log \mathcal{N}(\mathcal{V}, \rho_{\mathrm{Fin}}, t))^{1/2}\, dt\right]^2 + \widetilde{\alpha}^2 \log^2 \mathcal{N}(\mathcal{V}, \rho_{\mathrm{Fin}}, \varepsilon_0) + \varepsilon_0^2 p_{\mathcal{V}} \log \frac{1}{\varepsilon_0}\right), \tag{11}$$

where $\widetilde{\alpha}^2$ is the largest leverage score of any $\left[A^{\left\{\theta^{(r)}_{\backslash 1}\right\}}, A^{\left\{\phi^{(r)}_{\backslash 1}\right\}}\right] \in \mathcal{V}$ and $\mathcal{N}(\mathcal{V}, \rho_{\mathrm{Fin}}, t)$ is the covering number of $\mathcal{V}$ with radius $t$ under the Finsler metric.

Theorem 1 is based on recent work on a unified theory of dimensionality reduction (Dirksen, 2015; Bourgain et al., 2015). Note that the parameter space for the tensor regression problem (1) is a subspace of $\mathbb{R}^{\prod p_d}$, i.e., $\mathcal{S}_{\odot D,R} \subset \mathbb{R}^{\prod p_d}$. Therefore, a naïve application of sketching requires $m \gtrsim \prod p_d/\varepsilon^2$ in order for (8) to hold (Nelson and Nguyen, 2014). However, $\prod p_d$ can be very large and is far larger than the intrinsic number of degrees of freedom of the parameter space $\mathcal{S}_{\odot D,R}$, which is $R\sum p_d$. In the sequel, we provide a careful analysis of dimensionality reduction in terms of $\gamma_2(\mathcal{V},\rho_{\mathrm{Fin}})$, $p_{\mathcal{V}}$, and $\mathcal{N}(\mathcal{V},\rho_{\mathrm{Fin}},\eta_0)$, where sufficient conditions $m = \Omega(R\sum p_d)$ and $s = \Omega(1)$ are achieved, up to logarithmic factors.

### 3.1 Base Case: Rank-1 and Two-Way Tensors

We start with the base case when $R = 1$ and $D = 2$, i.e., the parameter space is $\mathcal{S}_{2,1}$. Then the parameter admits the decomposition $\Theta = \theta_1 \circ \theta_2$. For notational convenience, we let $\Theta = u \circ v$, where $u \in \mathbb{R}^{p_1}$ and $v \in \mathbb{R}^{p_2}$, and let $A^v = \sum_{i=1}^{p_2} A^{(i)} v_i$, where $A = [A^{(1)},\ldots,A^{(p_2)}] \in \mathbb{R}^{n\times p_2 p_1}$ with $A^{(i)} \in \mathbb{R}^{n\times p_1}$ for all $i \in [p_2]$. Consequently, the observation model (4) can be written as

$$b = A(v \otimes u) + z = A^v u + z,$$

and the corresponding OLS and SOLS using an SJLT matrix $\Phi \in \mathbb{R}^{m\times n}$ are, respectively,

$$\min_{v\in\mathbb{R}^{p_2},u\in\mathbb{R}^{p_1}} \|A^v u - b\|_2^2 \quad \text{and} \quad \min_{v\in\mathbb{R}^{p_2},u\in\mathbb{R}^{p_1}} \|\Phi A^v u - \Phi b\|_2^2.$$

Next, we show the following theorem, which provides sufficient conditions for the base case $\mathcal{S}_{2,1}$.

**Theorem 2.** Suppose the leverage scores of $A$ are bounded, i.e., $\max_{i\in[n]} \ell_i^2(A) \le 1/p_2^2$. Let

$$\mathcal{T} = \left\{ \frac{Ax - Ay}{\|Ax - Ay\|_2} \;\middle|\; x = v_1 \otimes u_1, y = v_2 \otimes u_2,\ u_1, u_2 \in \mathbb{R}^{p_1} \right\}$$

and $\Phi \in \mathbb{R}^{m\times n}$ is an SJLT matrix with column sparsity $s$. Then (8) holds if $m$ and $s$ satisfy

$$m \gtrsim \varepsilon^{-2}(p_1 + p_2)\log((p_1 + p_2)\kappa(A))(\log^4 m)(\log^5 n), \tag{12}$$
$$s \gtrsim \varepsilon^{-2}\log^2(p_1 + p_2)(\log^6 m)(\log^5 n). \tag{13}$$

From Theorem 2, when $m = \Omega(p_1 + p_2)$ and $s = \Omega(1)$, (9) holds.

### 3.2 Extension to General Ranks

We extend our analysis to the general case of two-way tensors with general rank, i.e., the parameter space is $\mathcal{S}_{2,R}$ for $R \ge 1$. In this case, we have $\Theta = \sum_{r=1}^{R} u^{(r)} \circ v^{(r)}$, where $u^{(r)} \in \mathbb{R}^{p_1}$ and $v^{(r)} \in \mathbb{R}^{p_2}$ for all $r \in [R]$, and $A^{\{v^{(r)}\}} = \left[\sum_{i=1}^{p_2} A^{(i)} v_i^{(1)},\ldots,\sum_{i=1}^{p_2} A^{(i)} v_i^{(R)}\right]$, where $A = [A^{(1)},\ldots,A^{(p_2)}] \in \mathbb{R}^{n\times p_2 p_1}$ and $A^{(i)} \in \mathbb{R}^{n\times p_1}$ for all $i \in [p_2]$. Consequently, the observation model (4) can be written as

$$b = A^{\{v^{(r)}\}}\left[u^{(1)\top} \ldots u^{(R)\top}\right]^\top + z,$$

and the corresponding OLS and SOLS using an SJLT matrix $\Phi \in \mathbb{R}^{m\times n}$ are, respectively,

$$\min_{\substack{v^{(r)}\in\mathbb{R}^{p_2}\\ u^{(r)}\in\mathbb{R}^{p_1},\forall r\in[R]}} \left\|A^{\{v^{(r)}\}}\left[u^{(1)\top}\dots u^{(R)\top}\right]^\top - b\right\|_2^2, \text{ and } \min_{\substack{v^{(r)}\in\mathbb{R}^{p_2}\\ u^{(r)}\in\mathbb{R}^{p_1},\forall r\in[R]}} \left\|\Phi A^{\{v^{(r)}\}}\left[u^{(1)\top}\dots u^{(R)\top}\right]^\top - \Phi b\right\|_2^2.$$

Our next theorem provides sufficient conditions for $\mathcal{S}_{2,R}$.

**Theorem 3.** Suppose $R \le p_2/2$ and the leverage scores of $A$ are bounded, i.e., $\max_{i\in[n]} \ell_i^2(A) \le 1/(R^2 p_2^2)$. Let

$$\mathcal{T} = \left\{ \frac{Ax - Ay}{\|Ax - Ay\|_2} \;\middle|\; x = \sum_{r=1}^{R} v_1^{(r)} \otimes u_1^{(r)}, y = \sum_{r=1}^{R} v_2^{(r)} \otimes u_2^{(r)},\ u_1^{(r)}, u_1^{(r)} \in \mathbb{R}^{p_1},\ \forall r\in[R]\right\}$$

and $\Phi \in \mathbb{R}^{m\times n}$ is an SJLT matrix with column sparsity $s$. Then (8) holds if $m$ and $s$ satisfy

$$\begin{aligned} m &\gtrsim \varepsilon^{-2}(\log^4 m)(\log^5 n) R(p_1+p_2)\log\left(R(p_1+p_2)\kappa(A)\right), \\ s &\gtrsim \varepsilon^{-2}(\log^6 m)(\log^5 n)\log^2\left(R(p_1+p_2)\kappa(A)\right). \end{aligned}$$

From Theorem 3, we have that when $m = \Omega(R(p_1+p_2))$ and $s = \Omega(1)$, (9) holds using an SJLT matrix $\Phi$. The extra condition of $R \le p_2/2$ is not restrictive, as in applications of low-rank tensors, typically $R \ll \min_{d\in[D]} p_d$.

### 3.3 Extension to General Tensors

We first extend our analysis to general tensors with rank 1, i.e., the parameter space is now $\mathcal{S}_{D,1}$ for $D \ge 2$. In this case, we have $\Theta = \theta_1 \circ \cdots \circ \theta_D$, where $\theta_d \in \mathbb{R}^{p_d}$ for all $d \in [D]$. Consequently, the observation model (4) can be written as

$$b = A\cdot(\theta_D \otimes \cdots \otimes \theta_1) + z = A^{\{\theta_{\setminus 1}\}}\cdot \theta_1 + z,$$

and the corresponding OLS and SOLS using an SJLT matrix $\Phi \in \mathbb{R}^{m\times n}$ are, respectively,

$$\min_{\substack{\theta_i\in\mathbb{R}^{p_i}\\ \forall i\in[D]}} \left\|A^{\{\theta_{\setminus 1}\}}\theta_1 - b\right\|_2^2 \text{ and } \min_{\substack{\theta_i\in\mathbb{R}^{p_i}\\ \forall i\in[D]}} \left\|\Phi A^{\{\theta_{\setminus 1}\}}\theta_1 - \Phi b\right\|_2^2.$$

Our next theorem provides sufficient conditions for $\mathcal{S}_{D,1}$.

**Theorem 4.** Suppose the leverage scores of $A$ are bounded, i.e., $\max_{i\in[n]} \ell_i^2(A) \le 1/\left(\sum_{d=2}^{D} p_d\right)^2$. For any $\vartheta = \theta_D \otimes \cdots \otimes \theta_1 \in \mathcal{S}_{\odot D,1}$ and $\varphi = \phi_D \otimes \cdots \otimes \phi_1 \in \mathcal{S}_{\odot D,1}$, $\theta_d, \phi_d \in \mathbb{R}^{p_d}$ for all $d \in [D]$, let

$$\mathcal{T} = \left\{ \frac{A\vartheta - A\varphi}{\|A\vartheta - A\varphi\|_2} \;\middle|\; \vartheta = \theta_D \otimes \cdots \otimes \theta_1, \varphi = \phi_D \otimes \cdots \otimes \phi_1,\ \theta_d, \phi_d \in \mathbb{R}^{p_d},\ \forall d\in[D]\right\}$$

and $\Phi \in \mathbb{R}^{m\times n}$ is an SJLT matrix with column sparsity $s$. Then (8) holds if $m$ and $s$ satisfy

$$\begin{aligned} m &\gtrsim \varepsilon^{-2}(\log^4 m)(\log^5 n)\left(\sum_{d=1}^{D} p_d \log\left(D\kappa(A)\sum_{d=1}^{D} p_d\right)\right), \\ s &\gtrsim \varepsilon^{-2}(\log^6 m)(\log^5 n)\log^2\left(\sum_{d=1}^{D} p_d\right). \end{aligned}$$

From Theorem 4, we have that when $m = \Omega\left(\sum_{d=1}^D p_d\right)$ and $s = \Omega(1)$, (9) holds using an SJLT matrix $\Phi$.

### 3.4 Extension to General Ranks and Tensors

Finally, we provide our guarantees for general tensors with general ranks, i.e., the parameter space is $\mathcal{S}_{D,R}$ for $D \geq 2$ and $R \geq 1$. We have the observation model (4) as

$$b = A \cdot \sum_{r=1}^{R} \theta_D^{(r)} \otimes \cdots \otimes \theta_1^{(r)} + z = \sum_{r=1}^{R} A^{\theta_{\setminus 1}^{(r)}} \cdot \theta_1^{(r)} + z = A^{\left\{\theta_{\setminus 1}^{(r)}\right\}} \cdot \left[\theta_1^{(1)\top} \dots \theta_1^{(R)\top}\right]^\top + z,$$

and the corresponding OLS and SOLS using an SJLT matrix $\Phi \in \mathbb{R}^{m\times n}$ are, respectively,

$$\min_{\substack{\theta_i^{(r)} \in \mathbb{R}^{p_i} \\ \forall i \in [D], r\in[R]}} \left\| A^{\left\{\theta_{\setminus 1}^{(r)}\right\}} \cdot \left[\theta_1^{(1)\top} \dots \theta_1^{(R)\top}\right]^\top - b \right\|_2^2, \text{ and } \min_{\substack{\theta_i^{(r)} \in \mathbb{R}^{p_i} \\ \forall i \in [D], r\in[R]}} \left\| \Phi A^{\left\{\theta_{\setminus 1}^{(r)}\right\}} \cdot \left[\theta_1^{(1)\top} \dots \theta_1^{(R)\top}\right]^\top - \Phi b \right\|_2^2.$$

Our most general theorem for CP decomposition is the following, providing sufficient conditions for $\mathcal{S}_{D,R}$.

**Theorem 5.** Suppose $R \leq \max_d p_d/2$ and the leverage scores of $A$ are bounded, i.e., $\max_{i\in[n]} \ell_i^2(A) \leq 1/\left(R^2\left(\sum_{d=2}^D p_d\right)^2\right)$. Let

$$\mathcal{T} = \left\{ \frac{A\vartheta - A\varphi}{\|A\vartheta - A\varphi\|_2} : \vartheta = \sum_{r=1}^{R} \theta_D^{(r)} \otimes \cdots \otimes \theta_1^{(r)}, \varphi = \sum_{r=1}^{R} \phi_D^{(r)} \otimes \cdots \otimes \phi_1^{(r)},\ \theta_d^{(r)}, \phi_d^{(r)} \in \mathcal{B}_{p_d},\ \forall r \in [R], d \in [D] \right\}$$

and $\Phi \in \mathbb{R}^{m\times n}$ is an SJLT matrix with column sparsity $s$. Then (8) holds if $m$ and $s$ satisfy

$$m \gtrsim \varepsilon^{-2} (\log^4 m)(\log^5 n) R \sum_{d=1}^{D} p_d \log\left(DR\kappa(A) \sum_{d=1}^{D} p_d\right),$$

$$s \gtrsim \varepsilon^{-2} (\log^6 m)(\log^5 n) \log^2\left(\sum_{d=1}^{D} p_d\right).$$

From Theorem 5, we have that when $m = \Omega\left(R\sum_{d=1}^D p_d\right)$ and $s = \Omega(1)$, (9) holds using an SJLT matrix $\Phi$. These complexities are optimal, up to logarithmic factors, for the CP decomposition model, since they meet the number of degrees of freedom of the CP model. The extra condition of $R \leq \max_d p_d/2$ is not restrictive, as we are interested in low-rank tensors satisfying $R \ll \min_{d\in[D]} p_d$.

## 4 Dimensionality Reduction for Tucker Decomposition

We start with a formal description of the Tucker model. Suppose $\Theta$ admits the following Tucker decomposition:

$$\Theta = \sum_{r_1=1}^{R_1} \cdots \sum_{r_D=1}^{R_D} g(r_1, \dots, r_D) \cdot \theta_1^{(r_1)} \circ \cdots \circ \theta_D^{(r_D)}, \tag{14}$$

where $\theta_d^{(r_d)} \in \mathbb{R}^{p_d}$ for all $r_d \in [R_d]$. Letting

$$A^{\theta_{\setminus 1}^{(r_1,\dots,r_D)}} = \sum_{j_D=1}^{p_D} \cdots \sum_{j_2=1}^{p_2} A^{(j_D,\dots,j_2)} \cdot \theta_{D,j_D}^{(r_D)} \cdots \theta_{2,j_2}^{(r_2)},$$

$$A^{\left\{\theta_{\setminus 1}^{\{r_d\}}\right\}} = \left[\sum_{r_2=1}^{R_2} \cdots \sum_{r_D=1}^{R_D} A^{\theta_{\setminus 1}^{(r_1,\dots,r_D)}} \cdot g(1, r_2, \dots, r_D), \dots, \sum_{r_2=1}^{R_2} \cdots \sum_{r_D=1}^{R_D} A^{\theta_{\setminus 1}^{(r_1,\dots,r_D)}} \cdot g(R_1, r_2, \dots, r_D)\right],$$

then the observation model (4) can be written as

$$b = A \sum_{r_1=1}^{R_1} \cdots \sum_{r_D=1}^{R_D} g(r_1, \dots, r_D) \cdot \theta_D^{(r_D)} \otimes \cdots \otimes \theta_1^{(r_1)} + z = \sum_{r_1=1}^{R_1} \cdots \sum_{r_D=1}^{R_D} A^{\theta_{\setminus 1}^{(r_1,\dots,r_D)}} \cdot g(r_1, \dots, r_D) \cdot \theta_1^{(r_1)} + z$$
$$= A^{\left\{\theta_{\setminus 1}^{\{r_d\}}\right\}} \cdot \left[\theta_1^{(1)\top} \dots \theta_1^{(R_1)\top}\right]^\top + z,$$

and the corresponding OLS and SOLS using an SJLT matrix $\Phi \in \mathbb{R}^{m \times n}$ are, respectively,

$$\min_{\substack{\theta_i^{(r)} \in \mathbb{R}^{p_i} \\ \forall i \in [D], r \in [R]}} \left\| A^{\left\{\theta_{\setminus 1}^{(r)}\right\}} \cdot \left[\theta_1^{(1)\top} \dots \theta_1^{(R)\top}\right]^\top - b \right\|_2^2, \text{ and } \min_{\substack{\theta_i^{(r)} \in \mathbb{R}^{p_i} \\ \forall i \in [D], r \in [R]}} \left\| \Phi A^{\left\{\theta_{\setminus 1}^{(r)}\right\}} \cdot \left[\theta_1^{(1)\top} \dots \theta_1^{(R)\top}\right]^\top - \Phi b \right\|_2^2.$$

Our next theorem provides sufficient conditions for the general Tucker decomposition model.

**Theorem 6.** Suppose $\prod_{d=1}^D R_d \le \max_d p_d/2$ and the leverage scores of $A$ are bounded, i.e., $\max_{i\in[n]} \ell_i^2(A) \le 1/\left(\sum_{d=2}^D R_d p_d + \prod_{d=1}^D p_d\right)^2$. Let

$$\mathcal{T} = \left\{ \frac{A\vartheta - A\varphi}{\|A\vartheta - A\varphi\|_2} : \vartheta = \sum_{r_1=1}^{R_1} \cdots \sum_{r_D=1}^{R_D} g_1(r_1, \dots, r_D) \cdot \theta_D^{(r_D)} \otimes \cdots \otimes \theta_1^{(r_1)}, \right.$$
$$\left. \varphi = \sum_{r_1=1}^{R_1} \cdots \sum_{r_D=1}^{R_D} g_2(r_1, \dots, r_D) \cdot \phi_D^{(r_D)} \otimes \cdots \otimes \phi_1^{(r_1)},\ \theta_d^{(r_d)}, \phi_d^{(r_d)} \in \mathcal{B}_{p_d},\ \forall r_d \in [R_d], d \in [D] \right\}$$

and $\Phi \in \mathbb{R}^{m \times n}$ is an SJLT matrix with column sparsity $s$. Then (8) holds if $m$ and $s$ satisfy

$$m \gtrsim \varepsilon^{-2} (\log^4 m)(\log^5 n) C_0 \cdot \log\left(C_0 D \kappa(A) \sqrt{\prod_{d=2}^D R_d}\right),$$
$$s \gtrsim \varepsilon^{-2} (\log^6 m)(\log^5 n) \log^2 C_0,$$

where $C_0 = \sum_{d=1}^D R_d p_d + \prod_{d=1}^D p_d$.

From Theorem 6, we have that when $m = \Omega\left(D(\sum_{d=1}^D R_d p_d + \prod_{d=1}^D p_d)\right)$ and $s = \Omega(D)$, then (8) holds for the Tucker decomposition model using an SJLT matrix, provided that $\prod R_d$ is not too large compared with $\max_d p_d$, which is typical in applications of low rank tensors in which the goal is to use small values of the $R_d$ when faced with large values of the $p_d$. Thus, the solution to the SOLS is a $(1+\epsilon)$-approximation to the OLS.

## 5 Flattening Leverage Scores

The analysis above depends on a bound on the leverage scores of the design matrix $A$. This might be restrictive if we have no control on the design $A$. In the sequel, we apply a standard idea (Tropp, 2011; Halko et al., 2011) to flatten the leverage scores of a deterministic design $A$ based on the Walsh-Hadamard matrix. An SRHT matrix is defined as

$$\Phi = \sqrt{\frac{n}{m}} PH\Sigma, \tag{15}$$

where the components $\Sigma$, $H$ and $P$ are generated as:

(G1) $\Sigma$ is an $n \times n$ diagonal matrix, where $\Sigma_{ii} = 1$ or -1 with equal probabilities 1/2.

(G2) $H$ is an $n \times n$ orthogonal matrix generated from a Walsh-Hadamard matrix scaled by $n^{-1/2}$.

(G3) $P$ is an $m \times n$ SJLT matrix, with column sparsity bounded by $s$.

Note that computing a matrix-vector product with $H$ takes $\mathcal{O}(n \log n)$ instead of $n^2$ time. Thus, one can compute $H\Sigma A$ for an $n \times d$ matrix $A$ in $O(nd \log n)$ time, which is well-suited for the case in which $A$ is dense, e.g., $\mathrm{nnz}(A) = \Theta(nd)$. The purpose of the matrix product $H\Sigma$ is to uniformize the leverage scores before applying our SJLT $P$.

We next give a standard lemma for flattening the leverage scores, included for completeness. Without loss of generality, we assume that $n = 2^q$ for a positive integer $q$, implying that a Walsh-Hadamard matrix exists.

**Lemma 1.** Suppose $H$ and $\Sigma$ are generated as in (G1) and (G2). Given any real value $\delta \in (0,1)$ and an $n \times d$ matrix $A$ with $\mathrm{rank}(A) = r$, we have with probability at least $1 - \delta$,

$$\max_{i \in [n]} \ell_i^2(H\Sigma A) \lesssim \frac{r \cdot \log\left(\frac{nr}{\delta}\right)}{n}.$$

Applying this with the bound $\max_{i \in [n]} \ell_i^2(H\Sigma A) \leq 1/\left(R^2 \left(\sum_{d=2}^D p_d\right)^2\right)$ of Theorem 5 gives:

**Proposition 1.** Suppose $H$ and $\Sigma$ are generated as in (G1) and (G2). For low-rank tensor regression (4), where $A \in \mathbb{R}^{n \times \prod_{d=1}^D p_d}$ is the matricization of all tensor designs, if $n$ satisfies

$$n \gtrsim R^2 \left(\sum_{d=2}^D p_d\right)^2 \cdot \mathrm{rank}(A) \cdot \log\left(\frac{n \cdot \mathrm{rank}(A)}{\delta}\right),$$

then with probability at least $1 - \delta$, we have

$$\max_{i \in [n]} \ell_i^2(H\Sigma A) \leq 1/\left(R^2 \left(\sum_{d=2}^D p_d\right)^2\right).$$

Combining Theorem 5 and Proposition 1, we achieve (8), provided $n$ is sufficiently large. Here we use that for all $x$, $\|H\Sigma Ax\|_2 = \|Ax\|_2$ since $H\Sigma$ is an isometry.

In the worst case, $\mathrm{rank}(A) = \prod_{d=1}^{D} p_d$, which requires $n = \Omega\left(R^2\left(\sum_{d=2}^{D} p_d\right)^2 \cdot \prod_{d=1}^{D} p_d\right)$. In over-constrained regression, it is often assumed that the number $n$ of examples is at least a small polynomial in $\mathrm{rank}(A)$ Woodruff (2014), which implies this bound on $n$. Also, if, for example, $A_i$ is sampled from a distribution with a rank deficient covariance, one may even have $\mathrm{rank}(A) \ll \prod_{d=1}^{D} p_d$.

One should note that computing $PH\Sigma A$ takes $(n \log n)\prod_{d=1}^{D} p_d$ time, provided the column sparsity $s$ of $P$ is $O(1)$. This is $O(\mathrm{nnz}(A)\log n)$ time for dense matrices $A$, i.e., those with $\mathrm{nnz}(A) = \Omega(nd)$, but in general, unlike our earlier results, is not $O(\mathrm{nnz}(A)\log n)$ time for sparse matrices. Analogous results can be obtained for the Tucker decomposition model, which we omit.

## 6 Experiments

We study the performance of sketching for tensor regression through numerical experiments over both synthetic and real data sets. For solving the OLS problem for tensor regression (2), we use a cyclic block-coordinate minimization algorithm based on a tensor toolbox (Zhou, 2013). Specifically, in a cyclic manner for all $d \in [D]$, we fix all but one $\Theta_d$ of $[[\Theta_1, \ldots, \Theta_D]] \in \mathcal{S}_{D,R}$ and minimize the resulting quadratic loss function (2) with respect to $\Theta_i$, until the decrease of the objective is smaller than a predefined threshold $\tau$. For SOLS, we use the same algorithm after multiplying $A$ and $b$ with an SJLT matrix $\Phi$. All results are run on a supercomputer due to the large scale of the data.

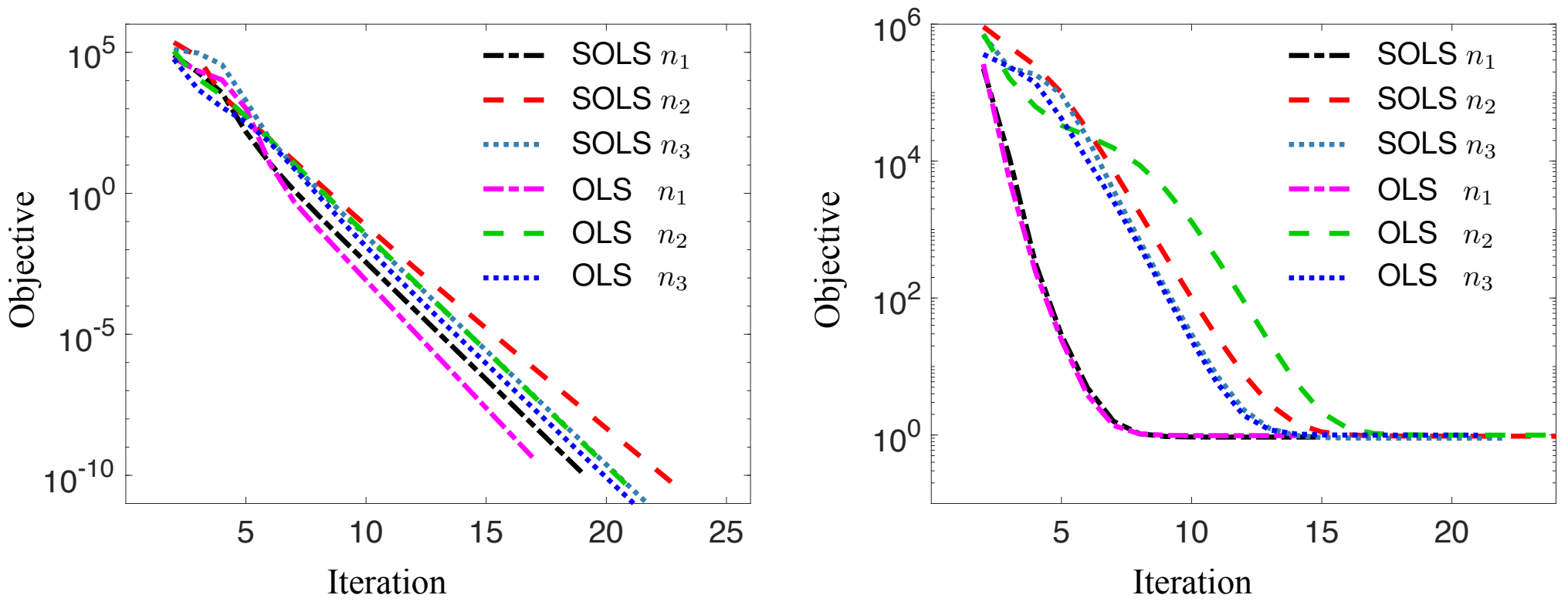

(a) $\sigma_z = 0$

(b) $\sigma_z = 1$

Figure 1: Comparison of SOLS and OLS for different settings on synthetic data. The vertical axis corresponds to the scaled objectives $\|A\vartheta^t_{\mathrm{SOLS}} - b\|_2^2/n$ for SOLS and $\|A\vartheta^t_{\mathrm{OLS}} - b\|_2^2/n$ for OLS, where $\vartheta^t$ is the update in the $t$-th iteration. The horizontal axis corresponds to the number of iterations (passes of block-coordinate minimization for all blocks). For both the noiseless case $\sigma_z = 0$ and noisy case $\sigma_z = 1$, we set $n_1 = 10^4$, $n_2 = 10^5$, and $n_3 = 10^6$ respectively.

For synthetic data, we generate the low-rank tensor $\Theta$ as follows. For every $d \in [D]$, we generate

Table 1: Comparison of SOLS and OLS on CPU execution time (in seconds) and the optimal scaled objective over different choices of sample sizes and noise levels on synthetic data. The results are averaged over 50 random trials, with both the mean values and standard deviations (in parentheses) provided. Note that we terminate the program after the running time exceeds $3\times10^4$ seconds.

| Variance of Noise | | $\sigma_z = 0$ | | | $\sigma_z = 1$ | | |
|---|---|---|---|---|---|---|---|
| Sample Size | | $n = 10^4$ | $n = 10^5$ | $n = 10^6$ | $n = 10^4$ | $n = 10^5$ | $n = 10^6$ |
| Time | OLS | 182.96 (72.357) | 3536.9 (1627.0) | $> 3 \times 10^4$ (NA) | 166.02 (5.6942) | 2620.4 (769.81) | $> 3 \times 10^4$ (NA) |
| | SOLS | 123.43 (37.452) | 132.81 (38.653) | 134.10 (36.406) | 122.641 (34.408) | 126.09 (35.719) | 127.98 (33.339) |
| Objective | OLS | $< 10^{-10}$ ($< 10^{-10}$) | $< 10^{-10}$ ($< 10^{-10}$) | $< 10^{-10}$ ($< 10^{-10}$) | 0.9089 (0.0217) | 0.9430 (0.0182) | 0.9440 (0.0137) |
| | SOLS | $< 10^{-10}$ ($< 10^{-10}$) | $< 10^{-10}$ ($< 10^{-10}$) | $< 10^{-10}$ ($< 10^{-10}$) | 0.9414 (0.0264) | 0.9854 (0.0227) | 0.9891 (0.0232) |

$R$ orthonormal columns to form $\Theta_d = [\theta_d^{(1)}, \ldots, \theta_d^{(R)}]$ of $[[\Theta_1, \ldots, \Theta_D]] \in \mathcal{S}_{D,R}$ independently. We also generate $R$ positive real scalars $\alpha_1, \ldots, \alpha_R$ uniformly and independently from $[1,10]$. Then $\Theta$ is formed by $\Theta = \sum_{r=1}^{R} \alpha_r \theta_1^{(r)} \circ \cdots \circ \theta_D^{(r)}$. The sequence of $n$ tensor designs are generated independently with i.i.d. $\mathcal{N}(0,1)$ entries for 10% of the entries chosen uniformly at random, and the remaining entries are set to zero. This allows for fast calculation of the leverage scores of matrix $A$, as well as memory savings. We also generate the noise $z$ to have i.i.d. $\mathcal{N}(0,\sigma_z^2)$ entries, and the generation of the SJLT matrix $\Phi$ follows Definition 3. For both OLS and SOLS, we use random initializations for $\Theta$, i.e., $\Theta_d$ has i.i.d. $\mathcal{N}(0,1)$ entries for all $d \in [D]$.

We compare OLS and SOLS for low-rank tensor regression under both the noiseless and noisy scenarios. For the noiseless case, i.e., $\sigma_z = 0$, we choose $R = 3$, $p_1 = p_2 = p_3 = 100$, $m = 5 \times R(p_1 + p_2 + p_3) = 4500$, and $s = 200$. Different values of $n \in \{10^4, 10^5, 10^6\}$ are chosen to compare both statistical and computational performances of OLS and SOLS. For the noisy case, the settings of all parameters are identical to those in the noiseless case, except that $\sigma_z = 1$. We provide a plot of the scaled objective versus the number of iterations for some random trials in Figure 1. The scaled objective is set as $\|A\vartheta_{\text{SOLS}}^t - b\|_2^2/n$ for SOLS and $\|A\vartheta_{\text{OLS}}^t - b\|_2^2/n$ for OLS, where $\vartheta_{\text{SOLS}}^t$ and $\vartheta_{\text{OLS}}^t$ are the updates in the $t$-th iterations of SOLS and OLS respectively. Note the we are using $\|\Phi A\vartheta_{\text{SOLS}} - \Phi b\|_2^2/n$ as the objective for solving the SOLS problem, but looking at the original objective $\|A\vartheta_{\text{SOLS}} - b\|_2^2/n$ for the solution of SOLS is ultimately what one is interested in. Moreover, the gap between $\|\Phi A\vartheta_{\text{SOLS}} - \Phi b\|_2^2/n$ and $\|A\vartheta_{\text{SOLS}} - b\|_2^2/n$ is very small in our results ($< 1\%$). The number of iterations is the number of passes of block-coordinate minimization for all blocks. We can see that OLS and SOLS require approximately the same number of iterations for comparable decrease of objective. However, since the SOLS instance has a much smaller size, its per iteration

Table 2: Comparison of SOLS and OLS on CPU execution time (in seconds) and the optimal scaled objective over different choices of ranks on the MRI data. The results are averaged over 10 random trials, with both the mean values and standard deviations (in parentheses) provided.

| Rank | | $R = 3$ | $R = 5$ | $R = 10$ |
|---|---|---|---|---|
| Time | OLS | 2824.4 (768.08) | 8137.2 (1616.3) | 26851 (8320.1) |
| | SOLS | 196.31 (68.180) | 364.09 (145.79) | 761.73 (356.76) |
| Objective | OLS | 16.003 (0.1378) | 11.164 (0.1152) | 6.8679 (0.0471) |
| | SOLS | 17.047 (0.1561) | 11.992 (0.1538) | 7.3968 (0.0975) |

computational cost is much lower than that of OLS.

We further provide numerical results on the running time (CPU execution time) and the optimal scaled objectives in Table 1. Using the same stopping criterion, we see that SOLS and OLS achieve comparable objectives (within $<5\%$ differences), matching our theory. In terms of the running time, SOLS is much faster than OLS, especially when $n$ is large. For example, when $n = 10^6$, SOLS is orders of magnitude faster than OLS while achieving a comparable objective function value. This matches our discussion on the computational cost of OLS and SOLS. Note that here we suppose the rank is known for our simulation, which can be restrictive in practice. We observe that if we choose a moderately larger rank than the true rank of the underlying model,then the result is similar to what we discussed above. Smaller values of the rank result in a much deteriorated statistical performance for both OLS and SOLS.

In addition, we examine sketching for tensor regression on a real dataset of MRI imaging (Rosset et al., 2004). The dataset consists of 56 frames of a human brain, each of which is of dimension $128 \times 128$ pixels, i.e., $p_1 = p_2 = 128$ and $p_3 = 56$. The generation of design tensors $\{A_i\}$ and linear measurements $b$ follows the same settings as for the synthetic data, with $\sigma_z = 0$. We choose three values of $R = 3, 5, 10$, and set $m = 5 \times R(p_1 + p_2 + p_3)$. The sample size is set to $n = 10^4$ for all settings of $R$. Analogous to the synthetic data, we provide numerical results for SOLS and OLS on the running time (CPU execution time) and the optimal scaled objectives. Again, we have that SOLS is much faster than OLS when they achieve comparable optimal objectives, under all settings of ranks.

# A Proof of Theorem 1

From the main result in (Bourgain et al., 2015), we have that (8) holds if $m$ and $s$ satisfy

$$
\begin{aligned}
m &\gtrsim \varepsilon^{-2}(\log^3 m)(\log^5 n)\gamma_2^2(\mathcal{V},\rho_{\mathrm{Fin}}) + \varepsilon^{-2}(\log^4 m)(\log^5 n)\left(p_{\mathcal{V}} + \log \mathcal{N}(\mathcal{V},\rho_{\mathrm{Fin}},\varepsilon_0)\right),\\
s &\gtrsim \varepsilon^{-2}(\log^4 m)(\log^5 n)\left(\widetilde{\alpha}^2 \log^2 \mathcal{N}(\mathcal{V},\rho_{\mathrm{Fin}},\varepsilon_0) + \varepsilon_0^2 p_{\mathcal{V}} \log\frac{1}{\varepsilon_0} + \left[\int_0^{\varepsilon_0} (\log \mathcal{N}(\mathcal{V},\rho_{\mathrm{Fin}},t))^{1/2} dt\right]^2\right)\\
&\quad + \varepsilon^{-2}(\log^6 m)(\log^4 n),
\end{aligned}
$$

which can be obtained from (10) and (11).

# B Proof of Theorem 2

We start with an illustration that the set $\mathcal{T}$ can be reparameterized to the following set with respect to tensors with orthogonal factors:

$$
\mathcal{T} = \bigcup_{E\in\mathcal{V}} \{x \in E \mid \|x\|_2 = 1\}, \text{ where } \mathcal{V} = \bigcup_{\widetilde{\mathcal{W}}} \{\mathrm{span}[A^{v_1}, A^{v_2}]\} \text{ and } \widetilde{\mathcal{W}} = \left\{v_1, v_2 \in \mathcal{B}_{p_2} \text{ with } \langle v_1, v_2\rangle = 0\right\}.
$$

Suppose $\langle v_1, v_2\rangle \neq 0$, then let $v_2 = \alpha v_1 + \beta z$ for some $\alpha, \beta \in \mathbb{R}$ and a unit vector $z \in \mathbb{R}^{p_2}$, where $\langle v_1, z\rangle = 0$. Then we have

$$
\begin{aligned}
\frac{Ax - Ay}{\|Ax - Ay\|_2} &= \frac{A^{v_1}u_1 - A^{v_2}u_2}{\|A^{v_1}u_1 - A^{v_2}u_2\|_2} = \frac{A^{v_1}u_1 - A^{\alpha v_1 + \beta z}u_2}{\|A^{v_1}u_1 - A^{\alpha v_1 + \beta z}u_2\|_2} = \frac{A^{v_1}u_1 - A^{\alpha v_1}u_2 - A^{\beta z}u_2}{\|A^{v_1}u_1 - A^{\alpha v_1}u_2 - A^{\beta z}u_2\|_2}\\
&= \frac{A^{v_1}(u_1 - \alpha u_2) - A^{z}(\beta u_2)}{\|A^{v_1}(u_1 - \alpha u_2) - A^{z}(\beta u_2)\|_2},
\end{aligned}
$$

which is equivalent to $\langle v_1, v_2\rangle = 0$ by reparameterizing $z$ as $v_2$.

Next, by Theorem 1, we need to upper bound $\rho_{\mathcal{V}}$, $\gamma_2^2(\mathcal{V},\rho_{\mathrm{Fin}})$, and $\mathcal{N}(\mathcal{V},\rho_{\mathrm{Fin}},\varepsilon_0)$. These will be addressed separately as follows.

**Part 1: Bound $p_{\mathcal{V}}$.** For notational convenience, we denote $A^{v_1,v_2} = [A^{v_1}, A^{v_2}]$. It is straightforward that

$$
p_{\mathcal{V}} = \sup_{v_1,v_2\in\mathcal{B}_{p_2},\langle v_1,v_2\rangle=0} \dim\{\mathrm{span}(A^{v_1,v_2})\} \leq 2p_1. \tag{16}
$$

**Part 2: Bound $\gamma_2^2(\mathcal{V},\rho_{\mathrm{Fin}})$.** By the definition of $\gamma_2$-functional in (7) for the Finsler metric, we have

$$
\gamma_2(\mathcal{V},\rho_{\mathrm{Fin}}) = \inf_{\{\mathcal{V}_k\}_{k=0}^{\infty}} \sup_{A^{v_1,v_2}\in\mathcal{V}} \sum_{k=0}^{\infty} 2^{k/2}\cdot\rho_{\mathrm{Fin}}(A^{v_1,v_2}, \overline{\mathcal{V}}_k),
$$

where $\overline{\mathcal{V}}_k$ is an $\varepsilon_k$-net of $\mathcal{V}_k$, i,e., for any $A^{v_1,v_2} \in \mathcal{V}$ there exist $\overline{v}_1, \overline{v}_2 \in \mathcal{B}_{p_2}$ with $\langle \overline{v}_1, \overline{v}_2\rangle = 0$, $\|v_1 - \overline{v}_1\|_2 \leq \eta_k$, and $\|v_2 - \overline{v}_2\|_2 \leq \eta_k$, such that $A^{\overline{v}_1,\overline{v}_2} \in \overline{\mathcal{V}}_k$ and $\rho_{\mathrm{Fin}}(A^{v_1,v_2}, A^{\overline{v}_1,\overline{v}_2}) \leq \varepsilon_k$.

From Lemma 6, we have $\rho_{\text{Fin}}(A^{v_1,v_2},\overline{\mathcal{V}}_k)\le 2\kappa(A)\eta_k$ for $\|v_1-\overline{v}_1\|_2\le\eta_k$ and $\|v_2-\overline{v}_2\|_2\le\eta_k$. On the other hand, we have that $\rho_{\text{Fin}}(A^{v_1,v_2},\overline{\mathcal{V}}_k)\le 1$ always holds. Therefore, we have

$$\rho_{\text{Fin}}(A^{v_1,v_2},\overline{\mathcal{V}}_k)\le\min\{2\kappa(A)\eta_k,1\}.$$

Let $k'$ be the smallest integer such that $2\kappa(A)\eta_{k'}\le 1$. Then we have

$$\gamma_2(\mathcal{V},\rho_{\text{Fin}})\le\sum_{k=0}^{\infty}2^{k/2}\cdot\rho_{\text{Fin}}(A^{v_1,v_2},\overline{\mathcal{V}}_k)\le\sum_{k=0}^{k'}2^{k/2}+\sum_{k=k'+1}^{\infty}2^{k/2}\cdot\rho_{\text{Fin}}(A^{v_1,v_2},\overline{\mathcal{V}}_k). \tag{17}$$

Suppose that $\eta_0=1$. Then we have $|\overline{\mathcal{V}}_0|=1$. For $k\ge 1$, we have $\eta_k<1$ and $|\overline{\mathcal{V}}_k|\le(3/\eta_k)^{p_2}$ (Vershynin, 2010). By the definition of admissible sequences in the $\gamma_2$-functional, we require $|\overline{\mathcal{V}}_k|\le 2^{2^k}$. Without loss of generality, suppose that for all $k\le k'$, we have $|\overline{\mathcal{V}}_k|\le 2^{2^k}\le(3/\eta_k)^{p_2}$. Then we have $2^{k/2}\le\sqrt{p_2\log\frac{3}{\eta_k}}$, which implies

$$\sum_{k=0}^{k'}2^{k/2}=\frac{2^{k'/2}}{\sqrt{2}-1}\lesssim\sqrt{p_2\log\frac{1}{\eta_{k'}}}. \tag{18}$$

For $k>k'$, suppose we choose $\eta_{k+1}=\eta_k^2$. Then we have

$$\left(\frac{3}{\eta_{k+1}}\right)^{p_2}\le\left(\frac{3}{\eta_k}\right)^{2p_2}\le\left(2^{2^k}\right)^2=2^{2^{k+1}}, \tag{19}$$

which implies $|\overline{\mathcal{V}}_{k+1}|\le 2^{2^{k+1}}$ as long as $|\overline{\mathcal{V}}_{k+1}|\le(3/\eta_{k+1})^{p_2}$ holds. In other words, we have $|\overline{\mathcal{V}}_k|\le 2^{2^k}$ if we choose $\eta_{k+1}=\eta_k^2$ for all $k>k'$. Suppose $k'$ is the smallest integer such that when we choose $\eta_{k'+1}=\frac{1}{4\kappa(A)}$, then $\left(\frac{3}{\eta_{k'+1}}\right)^{p_2}\le 2^{2^{k'+1}}$ holds. This implies (19) holds and $\rho_{\text{Fin}}(A^{v_1,v_2},\overline{\mathcal{V}}_k)\le(1/2)^{2^{k-k'}}$ for all $k>k'$. Then we have

$$\sum_{k=k'+1}^{\infty}2^{k/2}\cdot\rho_{\text{Fin}}(A^{v_1,v_2},\overline{\mathcal{V}}_k)=2^{k'/2}\cdot\sum_{t=1}^{\infty}2^{t/2}\cdot\left(\frac{1}{2}\right)^{2^t}\le 2^{k'/2}\lesssim\sqrt{p_2\log\frac{1}{\eta_{k'}}}, \tag{20}$$

where the first inequality is from the Cauchy condensation test $\sum_{t=0}^{\infty}2^{t/2}\cdot\left(\frac{1}{2}\right)^{2^t}\le 2\cdot\sum_{t=0}^{\infty}\left(\frac{1}{2}\right)^t=1$ and the second inequality is from (18).

Combining (17), (18), and (20), we have

$$\gamma_2^2(\mathcal{V},\rho_{\text{Fin}})\lesssim p_2\log\frac{1}{\eta_{k'}}. \tag{21}$$

From Lemma 6, suppose we choose a small enough $\varepsilon_0$ such that $\varepsilon_0\le 2\kappa(A)\eta_{k'}$. Then (21) implies

$$\gamma_2^2(\mathcal{V},\rho_{\text{Fin}})\lesssim p_2\log\frac{\kappa(A)}{\varepsilon_0}. \tag{22}$$

**Part 3: Bound $\mathcal{N}(\mathcal{V}, \rho_{\textbf{Fin}}, \varepsilon_0)$.** From our choice from Part 2, $\varepsilon_0 \in (0,1)$ is a constant. Then it is straightforward that

$$\mathcal{N}(\mathcal{V}, \rho_{\mathrm{Fin}}, \varepsilon_0) \leq \left(\frac{3}{\varepsilon_0}\right)^{2p_2}. \tag{23}$$

This implies

$$\begin{aligned}
&\int_0^{\varepsilon_0} [\log \mathcal{N}(\mathcal{V}, \rho_{\mathrm{Fin}}, t)]^{1/2} dt \leq \int_0^{\varepsilon_0} (\log (3/t)^{p_2})^{1/2}\, dt \lesssim \sqrt{p_2} \int_0^{\varepsilon_0} (-\log t)^{1/2}\, dt \quad \left(\text{Let } w = (-\log t)^{1/2}\right) \\
&= \sqrt{p_2} \int_{-\infty}^{(-\log \varepsilon_0)^{1/2}} 2w^2 e^{-w^2} dw = \sqrt{p_2}\left( \left[ w \cdot e^{-w^2}\right]_{-\infty}^{(-\log \varepsilon_0)^{1/2}} - \int_{-\infty}^{(-\log \varepsilon_0)^{1/2}} e^{-w^2} dw \right) \\
&\leq \sqrt{p_2} \left[ w \cdot e^{-w^2}\right]_{-\infty}^{(-\log \varepsilon_0)^{1/2}} = \varepsilon_0 \sqrt{p_2 \log \frac{1}{\varepsilon_0}}.
\end{aligned} \tag{24}$$

From Lemma 4, we have

$$\widetilde{\alpha}^2 = \max_{i \in [n]} \ell_i^2 (A^{v_1, v_2}) \leq \max_{i \in [n]} \ell_i^2(A) \leq 1/p_2^2. \tag{25}$$

Combining (16), (22)–(25), and Theorem 1, we have that the claim holds if

$$\begin{aligned}
m &\gtrsim \varepsilon^{-2} \left( p_2 \log \frac{\kappa(A)}{\varepsilon_0} + p_1 + p_2 \log \frac{1}{\varepsilon_0} \right) (\log^4 m)(\log^5 n), \\
s &\gtrsim \varepsilon^{-2} \left( \log^2 \frac{1}{\varepsilon_0} + \varepsilon_0^2 (p_1 + p_2) \log \frac{1}{\varepsilon_0} \right) (\log^6 m)(\log^5 n).
\end{aligned}$$

Taking $\varepsilon_0 = 1/(p_1+p_2)$, we finish the proof. Note that since $2\kappa(A)\eta_{k'} \geq 1/2$, we only require $\rho_{\mathrm{Fin}}(A^{v_1,v_2}, \overline{\mathcal{V}}_{k'}) \leq 1/2$ in Part 2. Thus the choice $\varepsilon_0 = 1/(p_1+p_2)$ is valid here.

# C Proof of Theorem 3

Denote $A^{\left\{v_i^{(r)}\right\}} = \left[A^{\left\{v_1^{(r)}\right\}}, A^{\left\{v_2^{(r)}\right\}}\right] \in \mathbb{R}^{n \times 2Rp_1}$. We illustrate that the set $\mathcal{T}$ can be reparameterized to the following set with respect to tensors with partial orthogonal factors:

$$\begin{aligned}
&\mathcal{T} = \bigcup_{E \in \mathcal{V}} \{x \in E \mid \|x\|_2 = 1\}, \;\; \text{where } \; \mathcal{V} = \bigcup_{\widetilde{\mathcal{W}}} \mathrm{span}\left(A^{\left\{v_i^{(r)}\right\}}\right) \;\text{ and} \\
&\widetilde{\mathcal{W}} = \left\{\forall i \in [2],\ r,q \in [R],\ q \neq r, v_i^{(r)} \in \mathcal{B}_{p_2}, \langle v_1^{(r)}, v_2^{(r)} \rangle = \langle v_i^{(r)}, v_i^{(q)} \rangle = 0\right\}.
\end{aligned}$$

Suppose for all $r \in [R]$, $v_2^{(r)} = \alpha^{(r)} v_1^{(r)} + \beta^{(r)} z^{(r)}$ for some $\alpha^{(r)}, \beta^{(r)} \in \mathbb{R}$ and unit vectors $z^{(r)} \in \mathbb{R}^{p_2}$, where $\langle v_1^{(r)}, z^{(r)} \rangle = 0$. Then we have

$$\begin{aligned}
Ax - Ay &= \sum_{r=1}^R \left( A^{v_1^{(r)}} \cdot u_1^{(r)} - A^{v_2^{(r)}} \cdot u_2^{(r)} \right) = \sum_{r=1}^R \left( A^{v_1^{(r)}} \cdot u_1^{(r)} - A^{\alpha^{(r)} v_1^{(r)} + \beta^{(r)} z^{(r)}} \cdot u_2^{(r)} \right) \\
&= \sum_{r=1}^R \left( A^{v_1^{(r)}} \cdot u_1^{(r)} - A^{\alpha^{(r)} v_1^{(r)}} \cdot u_2^{(r)} - A^{\beta^{(r)} z^{(r)}} \cdot u_2^{(r)} \right) = \sum_{r=1}^R \left( A^{v_1^{(r)}} \cdot \left( u_1^{(r)} - \alpha^{(r)} u_2^{(r)} \right) - A^{z^{(r)}} \cdot \left( \beta^{(r)} u_2^{(r)} \right) \right).
\end{aligned}$$

which is equivalent to $\langle v_1^{(r)}, v_2^{(r)}\rangle = 0$ by reparameterizing $z^{(r)}$ as $v_2^{(r)}$.

Using a similar argument, we show the general scenario. For any $r \in [R]$, $r \geq 2$, w.l.o.g., suppose

$$v_1^{(r)} = \alpha_1^{(r,1)} v_1^{(1)} + \sum_{i=2}^{r} \alpha_1^{(r,i)} z_1^{(i)} \ \text{ and } \ v_2^{(r)} = \beta_1^{(r,1)} v_1^{(1)} + \sum_{i=2}^{r} \beta_1^{(r,i)} z_1^{(i)} + \sum_{j=1}^{r} \beta_2^{(r,j)} z_2^{(j)}.$$

where $\alpha_1^{(r,i)}, \beta_1^{(r,i)}, \beta_2^{(r,j)} \in \mathbb{R}$ are real coefficients and $\langle v_1^{(1)}, z_1^{(i)}\rangle = \langle v_1^{(1)}, z_2^{(i)}\rangle = \langle z_1^{(i)}, z_2^{(j)}\rangle = 0$ for any $i, j \in [r]$. For $R = 1$, the argument is identical to the one above. For $2 \leq R \leq p_2/2$, we have

$$\begin{aligned} Ax - Ay &= \sum_{r=1}^{R} \left( A^{v_1^{(r)}} \cdot u_1^{(r)} - A^{v_2^{(r)}} \cdot u_2^{(r)} \right) \\ &= \sum_{r=2}^{R} \left( A^{\alpha_1^{(r,1)} v_1^{(1)} + \sum_{i=2}^{r} \alpha_1^{(r,i)} z_1^{(i)}} \cdot u_1^{(r)} - A^{\beta_1^{(r,1)} v_1^{(1)} + \sum_{i=2}^{r} \beta_1^{(r,i)} z_1^{(i)} + \sum_{j=1}^{r} \beta_2^{(r,j)} z_2^{(j)}} \cdot u_2^{(r)} \right) \\ &\qquad + A^{v_1^{(1)}} \cdot u_1^{(r)} - A^{\left(\beta_1^{(1,1)} v_1^{(1)} + \beta_2^{(1,1)} z_2^{(1)}\right)} \cdot u_2^{(r)} \\ &= \sum_{r=2}^{R} \left( A^{v_1^{(1)}} \cdot \left( \alpha_1^{(r,1)} u_1^{(1)} - \beta_1^{(r,1)} u_2^{(1)} \right) + \sum_{i=2}^{r} A^{z_1^{(i)}} \cdot \left( \alpha_1^{(r,i)} u_1^{(i)} - \beta_1^{(r,i)} u_2^{(i)} \right) - \sum_{j=1}^{r} A^{z_2^{(j)}} \cdot \left( \beta_2^{(r,j)} u_2^{(j)} \right) \right) \\ &\qquad + A^{v_1^{(1)}} \cdot u_1^{(r)} - A^{\left(\beta_1^{(1,1)} v_1^{(1)} + \beta_2^{(1,1)} z_2^{(1)}\right)} \cdot u_2^{(r)}, \end{aligned}$$

which is equivalent to $\langle v_i^{(r)}, v_i^{(q)}\rangle = 0$ and $\langle v_1^{(r)}, v_2^{(r)}\rangle = 0$ for all $i \in [2]$, $r \in [R]$, and $q \neq r$ by reparameterizing $z_1^{(i)}$ as $v_1^{(i)}$ and $z_2^{(j)}$ as $v_2^{(j)}$.

Next, analogous to Theorem 2, we analyze upper bounds on $\rho_{\mathcal{V}}$, $\gamma_2^2(\mathcal{V}, \rho_{\text{Fin}})$, and $\mathcal{N}(\mathcal{V}, \rho_{\text{Fin}}, \varepsilon_0)$, and obtain the result from Theorem 1.

**Part 1: Bound $p_{\mathcal{V}}$.** It is straightforward that

$$p_{\mathcal{V}} = \sup_{\widetilde{\mathcal{W}}} \dim\left\{ \operatorname{span}\left( A^{\left\{v_i^{(r)}\right\}} \right) \right\} \leq 2Rp_1. \tag{26}$$

**Part 2: Bound $\gamma_2^2(\mathcal{V}, \rho_{\text{Fin}})$.** The $\gamma_2$-functional in this case is

$$\gamma_2^2(\mathcal{V}, \rho_{\text{Fin}}) = \inf_{\{\mathcal{V}_k\}_{k=0}^{\infty}} \sup_{A^{\left\{v_i^{(r)}\right\}} \in \mathcal{V}} \sum_{k=0}^{\infty} 2^{r/2} \cdot \rho_{\text{Fin}}\left( A^{\left\{v_i^{(r)}\right\}}, \overline{\mathcal{V}}_k \right),$$

where $\overline{\mathcal{V}}_k$ is an $\varepsilon_k$-net of $\mathcal{V}_k$.

Following the same argument in Part 2 of the proof for Theorem 2, we have from Lemma 7 that if $k'$ is the smallest integer such that $2R\kappa(A)\eta_{k'} \leq 1$ and we choose $\eta_{k'+1} = \frac{1}{4R\kappa(A)}$, then we choose a small enough $\varepsilon_0$ such that $\varepsilon_0 \leq 2R\kappa(A)\eta_{k'}$,

$$\gamma_2^2(\mathcal{V}, \rho_{\text{Fin}}) \lesssim Rp_2 \log \frac{R\kappa(A)}{\varepsilon_0}. \tag{27}$$

**Part 3: Bound $\mathcal{N}(\mathcal{V}, \rho_{\textbf{Fin}}, \varepsilon_0)$.** It is straightforward that

$$\mathcal{N}(\mathcal{V}, \rho_{\text{Fin}}, \varepsilon_0) \leq \left(\frac{3}{\varepsilon_0}\right)^{2Rp_2}.$$

Following the same argument in Part 3 of the proof for Theorem 2, we have

$$\int_0^{\varepsilon_0} [\log \mathcal{N}(\mathcal{V}, \rho_{\text{Fin}}, t)]^{1/2} dt \lesssim \varepsilon_0 \sqrt{Rp_2 \log \frac{1}{\varepsilon_0}}. \tag{28}$$

From Lemma 5, we have

$$\widetilde{\alpha}^2 = \max_{i \in [n]} \ell_i^2 \left( A^{\left\{v_i^{(r)}\right\}} \right) \leq \max_{i \in [n]} \ell_i^2(A) \leq 1/(R^2 p_2^2). \tag{29}$$

Combining (26) – (29) and Theorem 1, we have that the claim holds if

$$m \gtrsim \varepsilon^{-2} R \left( p_2 \log \frac{R\kappa(A)}{\varepsilon_0} + p_1 + p_2 \log \frac{1}{\varepsilon_0} \right) (\log^4 m)(\log^5 n),$$
$$s \gtrsim \varepsilon^{-2} \left( \log^2 \frac{1}{\varepsilon_0} + \varepsilon_0^2 R(p_1 + p_2) \log \frac{1}{\varepsilon_0} \right) (\log^6 m)(\log^5 n).$$

We finish the proof by taking $\varepsilon_0 = \frac{1}{R(p_1+p_2)}$. Note that this choice of $\varepsilon$ satisfies the requirement in Part 2.

# D Proof of Theorem 4

Denote $\vartheta_{\backslash 1} = \theta_D \otimes \cdots \otimes \theta_2$, $\varphi_{\backslash 1} = \phi_D \otimes \cdots \otimes \phi_2$ and $A^{\vartheta_{\backslash 1}, \varphi_{\backslash 1}} = \left[ A^{\{\theta_{\backslash 1}\}}, A^{\{\phi_{\backslash 1}\}} \right] \in \mathbb{R}^{n \times 2p_1}$. We illustrate that the set $\mathcal{T}$ can be reparameterized to the following set with respect to tensors with partial orthogonal factors:

$$\mathcal{T} = \bigcup_{E \in \mathcal{V}} \{x \in E \mid \|x\|_2 = 1\}, \text{ where } \mathcal{V} = \bigcup_{\widetilde{\mathcal{W}}} \text{span}\left( A^{\vartheta_{\backslash 1}, \varphi_{\backslash 1}} \right) \text{ and}$$
$$\widetilde{\mathcal{W}} = \left\{ \forall d \in [D] \backslash \{1\}, \theta_d, \phi_d \in \mathcal{B}_{p_d}, \exists i \in [D] \backslash \{1\} \text{ s.t. } \langle \theta_i, \phi_i \rangle = 0 \right\},$$

W.l.o.g., suppose $\phi_D = \alpha \theta_D + \beta z$ for some $\alpha, \beta \in \mathbb{R}$ and a unit vector $z \in \mathbb{R}^{p_D}$, where $\langle \theta_D, z \rangle = 0$. Then we have

$$\begin{aligned}
A\vartheta - A\varphi = A^{\{\theta_{\backslash 1}\}} \theta_1 - A^{\{\phi_{\backslash 1}\}} \phi_1 &= A(\theta_D \otimes \cdots \otimes \theta_2 \otimes I_{p_1}) \theta_1 - A(\phi_D \otimes \cdots \otimes \phi_2 \otimes I_{p_1}) \phi_1 \\
&= A(\theta_D \otimes \cdots \otimes \theta_2 \otimes I_{p_1}) \theta_1 - A((\alpha \theta_D + \beta z) \otimes \phi_{D-1} \otimes \cdots \otimes \phi_2 \otimes I_{p_1}) \phi_1 \\
&= A(\theta_D \otimes \cdots \otimes \theta_2 \otimes I_{p_1}) \theta_1 - A(\alpha \theta_D \otimes \cdots \otimes \phi_2 \otimes I_{p_1}) \phi_1 - A(\beta z \otimes \cdots \otimes \phi_2 \otimes I_{p_1}) \phi_1 \\
&= A^{\theta_D} (\theta_{D-1} \otimes \cdots \otimes \theta_1 - \alpha \phi_{D-1} \otimes \cdots \otimes \phi_1) - A^z (\phi_{D-1} \otimes \cdots \otimes \phi_1),
\end{aligned}$$

This is equivalent to $\langle \theta_D, \phi_D \rangle = 0$ by reparameterizing $z$ as $\phi_D$.

Next, analogous to Theorem 2, we analyze upper bounds on $\rho_{\mathcal{V}}$, $\gamma_2^2(\mathcal{V}, \rho_{\text{Fin}})$, and $\mathcal{N}(\mathcal{V}, \rho_{\text{Fin}}, \varepsilon_0)$, and obtain the result from Theorem 1.

**Part 1: Bound** $p_{\mathcal{V}}$. It is straightforward that

$$p_{\mathcal{V}} = \sup_{\widetilde{\mathcal{W}}} \dim\left\{\operatorname{span}\left(A^{\vartheta_{\backslash 1}, \varphi_{\backslash 1}}\right)\right\} \leq 2p_1. \tag{30}$$

**Part 2: Bound** $\gamma_2^2(\mathcal{V}, \rho_{\mathbf{Fin}})$. The $\gamma_2$-functional in this case is

$$\gamma_2^2(\mathcal{V}, \rho_{\text{Fin}}) = \inf_{\{\mathcal{V}_k\}_{k=0}^{\infty}} \sup_{A^{\vartheta_{\backslash 1}, \varphi_{\backslash 1}} \in \mathcal{V}} \sum_{k=0}^{\infty} 2^{r/2} \cdot \rho_{\text{Fin}}\left(A^{\vartheta_{\backslash 1}, \varphi_{\backslash 1}}, \overline{\mathcal{V}}_k\right),$$

where $\overline{\mathcal{V}}_k$ is an $\varepsilon_k$-net of $\mathcal{V}_k$.

Following the same argument in Part 2 of the proof of Theorem 2, we have from Lemma 8 that if $k'$ is the smallest integer such that $2\kappa(A)\left((1+\eta_{k'})^D - 1\right) \leq 1$, then we choose $\varepsilon_0$ small enough such that

$$\varepsilon \leq 2\kappa(A) D \eta_{k'} \leq 2\kappa(A)\left((1+\eta_{k'})^D - 1\right).$$

where the second inequality is from the binomial expansion. Then we have

$$\gamma_2^2(\mathcal{V}, \rho_{\text{Fin}}) \lesssim \sum_{d=2}^{D} p_d \cdot \log \frac{D\kappa(A)}{\varepsilon_0}. \tag{31}$$

**Part 3: Bound** $\mathcal{N}(\mathcal{V}, \rho_{\mathbf{Fin}}, \varepsilon_0)$. It is straightforward that

$$\mathcal{N}(\mathcal{V}, \rho_{\text{Fin}}, \varepsilon_0) \leq \left(\frac{3}{\varepsilon_0}\right)^{2\sum_{d=2}^{D} p_d}.$$

Following the same argument in Part 3 of the proof for Theorem 2, we have

$$\int_0^{\varepsilon_0} [\log \mathcal{N}(\mathcal{V}, \rho_{\text{Fin}}, t)]^{1/2} dt \lesssim \varepsilon_0 \sqrt{\sum_{d=2}^{D} p_d \log \frac{1}{\varepsilon_0}}. \tag{32}$$

From Lemma 4, we have

$$\widetilde{\alpha}^2 = \max_{i \in [n]} \ell_i^2\left(A^{\vartheta_{\backslash 1}, \varphi_{\backslash 1}}\right) \leq \max_{i \in [n]} \ell_i^2(A) \leq 1 / \left(\sum_{d=2}^{D} p_d\right)^2. \tag{33}$$

Combining (30) – (33) and Theorem 1, we have that the claim holds if

$$m \gtrsim \varepsilon^{-2}\left(p_1 + \sum_{d=2}^{D} p_d \cdot \log \frac{D\kappa(A)}{\varepsilon_0}\right)(\log^4 m)(\log^5 n),$$

$$s \gtrsim \varepsilon^{-2}\left(\log^2 \frac{1}{\varepsilon_0} + \varepsilon_0^2 \sum_{d=1}^{D} p_d \log \frac{1}{\varepsilon_0}\right)(\log^6 m)(\log^5 n).$$

We finish the proof by taking $\varepsilon_0 = \frac{1}{\sum_{d=1}^{D} p_d}$. Note that this choice of $\varepsilon$ satisfies the requirement in Part 2.

# E Proof of Theorem 5

Denote $A^{\left\{\vartheta_{\backslash 1}^{(r)},\varphi_{\backslash 1}^{(r)}\right\}} = \left[A^{\left\{\theta_{\backslash 1}^{(r)}\right\}}, A^{\left\{\phi_{\backslash 1}^{(r)}\right\}}\right]$. We illustrate that the set $\mathcal{T}$ can be reparameterized to the following set with respect to tensors with partial orthogonal factors:

$$\mathcal{T} = \bigcup_{E \in \mathcal{V}} \{x \in E \mid \|x\|_2 = 1\}, \text{ where } \mathcal{V} = \bigcup_{\widetilde{\mathcal{W}}} \operatorname{span}\left(A^{\left\{\vartheta_{\backslash 1}^{(r)},\varphi_{\backslash 1}^{(r)}\right\}}\right),$$

$$\widetilde{\mathcal{W}} = \Big\{\forall r \in [R], d \in [D]\backslash\{1\}, \theta_d^{(r)}, \phi_d^{(r)} \in \mathcal{B}_{p_d}; \forall r, q \in [R], \exists i \in [D]\backslash\{1\} \text{ s.t. } \langle \theta_i^{(r)}, \phi_i^{(q)} \rangle = 0;$$
$$\forall r \in [R-1], q \in [R]\backslash[r], \exists j, k \in [D]\backslash\{1\} \text{ s.t. } \langle \theta_j^{(r)}, \theta_j^{(q)} \rangle = \langle \phi_k^{(r)}, \phi_k^{(q)} \rangle = 0\Big\}.$$

For $R = 1$, the argument is identical to the analysis in Theorem 4. For any $r \in [R]$, $r \geq 2$, w.l.o.g., suppose

$$\theta_D^{(r)} = \alpha_1^{(r,1)} \theta_D^{(1)} + \sum_{i=2}^{r} \alpha_1^{(r,i)} z_1^{(i)} \text{ and } \phi_D^{(r)} = \beta_1^{(r,1)} \theta_D^{(1)} + \sum_{i=2}^{r} \beta_1^{(r,i)} z_1^{(i)} + \sum_{j=1}^{r} \beta_2^{(r,j)} z_2^{(j)},$$

where $\alpha_1^{(r,i)}, \beta_1^{(r,i)}, \beta_2^{(r,j)} \in \mathbb{R}$ are real coefficients and $\langle \theta_D^{(1)}, z_1^{(i)} \rangle = \langle \theta_D^{(1)}, z_2^{(i)} \rangle = \langle z_1^{(i)}, z_2^{(j)} \rangle = 0$ for any $i, j \in [r]$. Then for $2 \leq R \leq p_2/2$, we have

$$
\begin{aligned}
&A\vartheta - A\varphi = A \cdot \sum_{r=1}^{R} \left(\theta_D^{(r)} \otimes \cdots \otimes \theta_2^{(r)} \otimes I_{p_1}\right) \theta_1^{(r)} - A \cdot \sum_{r=1}^{R} \left(\phi_D^{(r)} \otimes \cdots \otimes \phi_2^{(r)} \otimes I_{p_1}\right) \phi_1^{(r)} \\
&= A \cdot \sum_{r=2}^{R} \left( \left( \alpha_1^{(r,1)} \theta_D^{(1)} + \sum_{i=2}^{r} \alpha_1^{(r,i)} z_1^{(i)} \right) \otimes \cdots \otimes \theta_1^{(r)} \right) + A \cdot \left( \theta_D^{(1)} \otimes \cdots \otimes \theta_1^{(1)} \right) \\
&\quad - A \cdot \sum_{r=2}^{R} \left( \left( \beta_1^{(r,1)} \theta_D^{(1)} + \sum_{i=2}^{r} \beta_1^{(r,i)} z_1^{(i)} + \sum_{j=1}^{r} \beta_2^{(r,j)} z_2^{(j)} \right) \otimes \cdots \otimes \phi_1^{(r)} \right) - A \cdot \left( \left( \beta_1^{(1,1)} \theta_D^{(1)} + \beta_2^{(1,1)} z_2^{(1)} \right) \otimes \cdots \otimes \phi_1^{(1)} \right) \\
&= \sum_{r=r}^{R} A^{\theta_D^{(1)}} \left( \alpha_1^{(r,1)} \theta_{D-1}^{(r)} \otimes \cdots \otimes \theta_1^{(r)} - \beta_1^{(r,1)} \phi_{D-1}^{(r)} \otimes \cdots \otimes \phi_1^{(r)} \right) \\
&\quad + \sum_{r=2}^{R} \sum_{i=2}^{r} A^{z_1^{(1)}} \left( \alpha_1^{(r,i)} \theta_{D-1}^{(r)} \otimes \cdots \otimes \theta_1^{(r)} - \beta_1^{(r,i)} \phi_{D-1}^{(r)} \otimes \cdots \otimes \phi_1^{(r)} \right) - \sum_{r=1}^{R} \sum_{j=1}^{r} A^{z_2^{(j)}} \left( \beta_2^{(r,j)} \phi_{D-1}^{(r)} \otimes \cdots \otimes \phi_1^{(r)} \right)
\end{aligned}
$$

where $\alpha_1^{(,1)} = 1$. This is equivalent to $\langle \theta_D^{(r)}, \phi_D^{(r)} \rangle = 0$, $\langle \theta_D^{(r)}, \theta_D^{(q)} \rangle = 0$, and $\langle \phi_D^{(r)}, \phi_D^{(q)} \rangle = 0$ for all $r \in [R]$ and $q \neq [R]\backslash[r]$, by reparameterizing $z_1^{(i)}$ and $z_2^{(j)}$ as $\theta_D^{(i)}$ and $\phi_D^{(j)}$ properly. The remaining pairs of orthogonality in $\widetilde{\mathcal{W}}$ can be checked analogously by repeating the argument above.

**Part 1: Bound** $p_{\mathcal{V}}$. It is straightforward that

$$p_{\mathcal{V}} = \sup_{\widetilde{\mathcal{W}}} \dim \left\{ \operatorname{span}\left( A^{\left\{\vartheta_{\backslash 1}^{(r)},\varphi_{\backslash 1}^{(r)}\right\}} \right) \right\} \leq 2Rp_1. \tag{34}$$

**Part 2: Bound $\gamma_2^2(\mathcal{V}, \rho_{\mathbf{Fin}})$.** The $\gamma_2$-functional in this case is

$$\gamma_2^2(\mathcal{V}, \rho_{\text{Fin}}) = \inf_{\{\mathcal{V}_k\}_{k=0}^\infty} \sup_{A^{\left\{\vartheta_{\backslash 1}^{(r)}, \varphi_{\backslash 1}^{(r)}\right\}} \in \mathcal{V}} \sum_{k=0}^{\infty} 2^{r/2} \cdot \rho_{\text{Fin}} \left( A^{\left\{\vartheta_{\backslash 1}^{(r)}, \varphi_{\backslash 1}^{(r)}\right\}}, \overline{\mathcal{V}}_k \right),$$

where $\overline{\mathcal{V}}_k$ is an $\varepsilon_k$-net of $\mathcal{V}_k$.

Following the same argument in Part 2 of the proof for Theorem 2, we have from Lemma 9 that if $k'$ is the smallest integer such that $2R\kappa(A)\left((1+\eta_{k'})^D - 1\right) \leq 1$, then we choose $\varepsilon_0$ small enough such that

$$\varepsilon \leq 2RD\kappa(A)\eta_{k'} \leq 2R\kappa(A)\left((1+\eta_{k'})^D - 1\right),$$

where the second inequality follows from the binomial expansion. Then we have

$$\gamma_2^2(\mathcal{V}, \rho_{\text{Fin}}) \lesssim \sum_{d=2}^{D} p_d \cdot \log \frac{RD\kappa(A)}{\varepsilon_0}. \tag{35}$$

**Part 3: Bound $\mathcal{N}(\mathcal{V}, \rho_{\mathbf{Fin}}, \varepsilon_0)$.** It is straightforward that

$$\mathcal{N}(\mathcal{V}, \rho_{\text{Fin}}, \varepsilon_0) \leq \left(\frac{3}{\varepsilon_0}\right)^{2R\sum_{d=2}^{D} p_d}.$$

Following the same argument in Part 3 of the proof for Theorem 2, we have

$$\int_0^{\varepsilon_0} [\log \mathcal{N}(\mathcal{V}, \rho_{\text{Fin}}, t)]^{1/2} dt \lesssim \varepsilon_0 \sqrt{R \sum_{d=2}^{D} p_d \log \frac{1}{\varepsilon_0}}. \tag{36}$$

From Lemma 4, we have

$$\widetilde{\alpha}^2 = \max_{i \in [n]} \ell_i^2 \left(A^{\vartheta_{\backslash 1}, \varphi_{\backslash 1}}\right) \leq \max_{i \in [n]} \ell_i^2(A) \leq 1 / \left(R \sum_{d=2}^{D} p_d\right)^2. \tag{37}$$

Combining (34) – (37) and Theorem 1, we have that the claim holds if

$$m \gtrsim \varepsilon^{-2} R \left(p_1 + \sum_{d=2}^{D} p_d \cdot \log \frac{RD\kappa(A)}{\varepsilon_0}\right) (\log^4 m)(\log^5 n),$$

$$s \gtrsim \varepsilon^{-2} \left(\log^2 \frac{1}{\varepsilon_0} + \varepsilon_0^2 R \sum_{d=1}^{D} p_d \log \frac{1}{\varepsilon_0}\right) (\log^6 m)(\log^5 n).$$

We finish the proof by taking $\varepsilon_0 = \frac{1}{R\sum_{d=1}^{D} p_d}$. Note that this choice of $\varepsilon$ satisfies the requirement in Part 2.

# F Proof of Theorem 6

Denote $A^{\left\{\vartheta_{\backslash 1}^{\{r_d\}},\varphi_{\backslash 1}^{\{r_d\}}\right\}}=\left[A^{\left\{\theta_{\backslash 1}^{\{r_d\}}\right\}},A^{\left\{\phi_{\backslash 1}^{\{r_d\}}\right\}}\right]$. We illustrate that the set $\mathcal{T}$ can be reparameterized to the following set with respect to tensors with partial orthogonal factors:

$$\mathcal{T}=\bigcup_{E\in\mathcal{V}}\{x\in E\mid \|x\|_2=1\},\;\text{ where }\;\mathcal{V}=\bigcup_{\widetilde{\mathcal{W}}}\text{span}\left(A^{\left\{\vartheta_{\backslash 1}^{\{r_d\}},\varphi_{\backslash 1}^{\{r_d\}}\right\}}\right)\;\text{ and}$$

$$\widetilde{\mathcal{W}}=\Big\{\forall r_d\in[R_d],d\in[D]\backslash\{1\},\theta_d^{(r_d)},\phi_d^{(r_d)}\in\mathcal{B}_{p_d};\forall r_d,q_d\in[R_d],\exists d\in[D]\backslash\{1\}\text{ s.t. }\langle\theta_d^{(r_d)},\phi_d^{(q_d)}\rangle=0;$$
$$\forall r_d\in[R_d-1],q_d\in[R_d]\backslash[r_d],\exists d,t\in[D]\backslash\{1\}\text{ s.t. }\langle\theta_d^{(r_d)},\theta_d^{(q_d)}\rangle=\langle\phi_t^{(r_d)},\phi_t^{(q_d)}\rangle=0\Big\}.$$

Repeating the argument in the proof of Theorem 5, we have the equivalence of $\mathcal{T}$ and the set above.

**Part 1: Bound** $p_{\mathcal{V}}$. It is straightforward that

$$p_{\mathcal{V}}=\sup_{\widetilde{\mathcal{W}}}\dim\left\{\text{span}\left(A^{\left\{\vartheta_{\backslash 1}^{\{r_d\}},\varphi_{\backslash 1}^{\{r_d\}}\right\}}\right)\right\}\le 2R_1p_1. \tag{38}$$

**Part 2: Bound** $\gamma_2^2(\mathcal{V},\rho_{\mathbf{Fin}})$. The $\gamma_2$-functional in this case is

$$\gamma_2^2(\mathcal{V},\rho_{\text{Fin}})=\inf_{\{\mathcal{V}_k\}_{k=0}^{\infty}}\sup_{A^{\left\{\vartheta_{\backslash 1}^{\{r_d\}},\varphi_{\backslash 1}^{\{r_d\}}\right\}}\in\mathcal{V}}\sum_{k=0}^{\infty}2^{r/2}\cdot\rho_{\text{Fin}}\left(A^{\left\{\vartheta_{\backslash 1}^{\{r_d\}},\varphi_{\backslash 1}^{\{r_d\}}\right\}},\overline{\mathcal{V}}_k\right),$$

where $\overline{\mathcal{V}}_k$ is an $\varepsilon_k$-net of $\mathcal{V}_k$.

Following the same argument as in Part 2 of the proof for Theorem 2, we have from Lemma 10 that if $k'$ is the smallest integer such that $2\kappa(A)\big((1+\eta_{k'})^D-1\big)\sqrt{\prod_{d=2}^D R_d}\le 1$, then we choose $\varepsilon_0$ small enough such that

$$\varepsilon\le 2D\kappa(A)\eta_{k'}\sqrt{\prod_{d=2}^{D}R_d}\le 2\kappa(A)\big((1+\eta_{k'})^D-1\big)\sqrt{\prod_{d=2}^{D}R_d},$$

where the second inequality follows from the binomial theorem. Then we have

$$\gamma_2^2(\mathcal{V},\rho_{\text{Fin}})\lesssim\left(\sum_{d=2}^{D}R_dp_d+\prod_{d=1}^{D}p_d\right)\cdot\log\frac{D\kappa(A)\sqrt{\prod_{d=2}^D R_d}}{\varepsilon_0}. \tag{39}$$

**Part 3: Bound** $\mathcal{N}(\mathcal{V},\rho_{\mathbf{Fin}},\varepsilon_0)$. It is straightforward that

$$\mathcal{N}(\mathcal{V},\rho_{\text{Fin}},\varepsilon_0)\le\left(\frac{3}{\varepsilon_0}\right)^{2\left(\sum_{d=2}^D R_dp_d+\prod_{d=1}^D p_d\right)}.$$

Following the same argument in Part 3 of the proof for Theorem 2, we have

$$\int_0^{\varepsilon_0} [\log \mathcal{N}(\mathcal{V}, \rho_{\text{Fin}}, t)]^{1/2} dt \lesssim \varepsilon_0 \sqrt{\left(\sum_{d=2}^{D} R_d p_d + \prod_{d=1}^{D} p_d\right) \log \frac{1}{\varepsilon_0}}. \tag{40}$$

From Lemma 4, we have

$$\widetilde{\alpha}^2 = \max_{i\in[n]} \ell_i^2\left(A^{\vartheta_{\setminus 1}, \varphi_{\setminus 1}}\right) \le \max_{i\in[n]} \ell_i^2(A) \le 1 / \left(\sum_{d=2}^{D} R_d p_d + \prod_{d=1}^{D} p_d\right)^2. \tag{41}$$

Combining (34) – (37) and Theorem 1, we have that the claim holds if

$$m \gtrsim \varepsilon^{-2}\left(R_1 p_1 + \left(\sum_{d=2}^{D} R_d p_d + \prod_{d=1}^{D} p_d\right) \cdot \log \frac{D\kappa(A)\sqrt{\prod_{d=2}^{D} R_d}}{\varepsilon_0}\right)(\log^4 m)(\log^5 n),$$
$$s \gtrsim \varepsilon^{-2}\left(\log^2 \frac{1}{\varepsilon_0} + \varepsilon_0^2 \left(\sum_{d=1}^{D} R_d p_d + \prod_{d=1}^{D} p_d\right) \log \frac{1}{\varepsilon_0}\right)(\log^6 m)(\log^5 n).$$

We finish the proof by taking $\varepsilon_0 = \frac{1}{\sum_{d=1}^{D} R_d p_d + \prod_{d=1}^{D} p_d}$. Note that this choice of $\varepsilon$ satisfies the requirement in Part 2.

# G Proof of Lemma 1

Given a unit vector $y \in \mathbb{R}^n$, let $Z_{jk} = H_{jk}\Sigma_{kk} y_k$ for all $j \in [n]$. Then from the independence of $H_{jk}$ and $\Sigma_{kk}$, we have

$$\mathbb{E}(Z_{jk}) = \mathbb{E}(H_{jk}\Sigma_{kk} y_k) = \mathbb{E}(H_{jk}) \cdot \mathbb{E}(\Sigma_{kk}) \cdot y_k = 0,$$
$$\text{Var}(Z_{jk}) \le \mathbb{E}(H_{jk}^2 \Sigma_{kk}^2 y_k^2) = \mathbb{E}(H_{jk}^2) \cdot \mathbb{E}(\Sigma_{kk}^2) \cdot y_k^2 = \frac{y_k^2}{n}.$$

From the Azuma-Hoeffding inequality, for any $t > 0$ we have

$$\mathbb{P}\left(\left|\sum_{k=1}^{n} Z_{jk}\right| > t\right) \le 2\exp\left(-\frac{nt^2}{2\sum_{k=1}^{n} y_k^2}\right) = 2\exp\left(-\frac{nt^2}{2}\right).$$

By taking $t = \sqrt{\frac{2\log\left(\frac{2nr}{\delta}\right)}{n}}$, we have

$$\mathbb{P}\left(\left|\sum_{k=1}^{n} Z_{jk}\right| > \sqrt{\frac{2\log\left(\frac{2nr}{\delta}\right)}{n}}\right) \le 2\exp\left(\log\left(\frac{\delta}{2nr}\right)\right) = \frac{\delta}{nr}.$$

By a union bound, we have

$$\mathbb{P}\left(\left\|H\Sigma y\right\|_\infty > \sqrt{\frac{2\log\left(\frac{2nr}{\delta}\right)}{n}}\right) = \mathbb{P}\left(\max_{j\in[n]}\left|\sum_{k=1}^{n} Z_{jk}\right| > \sqrt{\frac{2\log\left(\frac{2nr}{\delta}\right)}{n}}\right) \le \frac{\delta}{r}.$$

Suppose $A = UQ$, where $U \in \mathbb{R}^{n\times r}$ has orthonormal columns. Then we have for all $i \in [n]$ and $k \in [r]$,

$$\ell_i^2(H\Sigma A) = \ell_i^2(H\Sigma U) \le r \cdot \left(e_i^\top H\Sigma U e_k\right)^2.$$

Using a union bound again, we finish the proof by

$$\mathbb{P}\left(\max_{i\in[n]} \ell_i^2(H\Sigma A) > \frac{2r\log\left(\frac{2nr}{\delta}\right)}{n}\right) \le \mathbb{P}\left(\max_{i\in[n]} r\cdot \left\|e_i^\top H\Sigma U e_k\right\|_\infty^2 > \frac{2r\log\left(\frac{2nr}{\delta}\right)}{n}\right) \le \delta.$$

## H Intermediate Results

Here we introduce all intermediate results applied in our main analysis.

**Lemma 2.** Suppose for $A = [A^{(1)}, A^{(2)}, \dots, A^{(m)}] \in \mathbb{R}^{n\times mp}$, each $A^{(i)} \in \mathbb{R}^{n\times p}$ is a column-wise submatrix of $A$. Given a vector $v \in \mathbb{R}^m$, we have

$$\left\|\sum_{i=1}^m A^{(i)} v_i\right\|_2 \le \|A\|_2 \|v\|_2.$$

*Proof.* This is an extension of the Cauchy-Schwartz inequality. We have $\sum_{i=1}^m A^{(i)} v_i = A(v\otimes I_p)$, where $\otimes$ is the Kronecker product. This implies

$$\left\|\sum_{i=1}^m A^{(i)} v_i\right\|_2 = \|A(v\otimes I_p)\|_2 \le \|A\|_2 \|v\otimes I_p\|_2 = \|A\|_2\|v\|_2.$$

□

**Lemma 3.** Given two sequences of unit vectors $\{\phi_i\}_{i=1}^n$ and $\{\psi_i\}_{i=1}^n$, where $\phi_i, \psi_i \in \mathbb{R}^{p_i}$ with $\|\phi_i - \psi_i\|_2 \le \varepsilon$ for all $i \in [n]$, we have

$$\|\phi_1\otimes\phi_2\otimes\cdots\otimes\phi_n - \psi_1\otimes\psi_2\otimes\cdots\otimes\psi_n\|_2 \le (1+\varepsilon)^n - 1.$$

*Proof.* Suppose for all $i\in[n]$, we have $\psi_i = \phi_1 + x_i$ for some vector $x_i \in \mathbb{R}^{p_i}$. Then we have

$$\begin{aligned}
&\|\phi_1\otimes\cdots\otimes\phi_n - \psi_1\otimes\cdots\otimes\psi_n\|_2 = \|\phi_1\otimes\cdots\otimes\phi_n - (\phi_1 + x_i)\otimes\cdots\otimes(\psi_n + x_n)\|_2 \\
&\le \sum_{i=1}^n \|\phi_1\otimes\cdots\otimes x_i\otimes\cdots\otimes\phi_n\|_2 + \sum_{i=1}^n\sum_{j=1, j\neq i}^n \|\phi_1\otimes\cdots\otimes x_i\otimes\cdots\otimes x_j\otimes\cdots\otimes\phi_n\|_2 + \cdots + \|x_1\otimes\cdots\otimes x_n\|_2 \\
&\le \binom{n}{1}\varepsilon + \binom{n}{2}\varepsilon^2 + \cdots + \binom{n}{n}\varepsilon^n = (1+\varepsilon)^n - 1,
\end{aligned}$$

where the last inequality is from the fact that $\|v\otimes u\|_2 = \|v\|_2\|u\|_2$ for any vectors $v$ and $u$. □

**Lemma 4.** Suppose that $A \in \mathbb{R}^{n\times\prod_{d=1}^2 p_d}$ has leverage scores $\ell_i^2(A)$ for all $i\in[n]$. Then for any $v_1, v_2 \in \mathbb{R}^{p_2}$, the leverage scores of $A^{v_1,v_2} = [A^{v_1}, A^{v_2}] \in \mathbb{R}^{n\times 2p_1}$ are bounded by $\ell_i^2(A^{v_1,v_2}) \le \ell_i^2(A)$.

*Proof.* Let $Z$ have orthonormal columns and have the same span as the column space of $A$. Then we have $\ell_i^2(A) = \|e_i^\top Z\|_2^2$ for all $i \in [n]$. Since the column space of $A^{v_1,v_2}$ is a subspace of the column space of $A$, we can always find a column sub-matrix $Z_1 \in \mathbb{R}^{n\times 2p_1}$ of $Z$ such that $Z_1$ spans the column space of $A^{v_1,v_2}$. Therefore, for each $i \in [n]$, we have

$$\ell_i^2(A^{v_1,v_2}) = \|e_i^\top Z_1\|_2^2 \le \|e_i^\top Z\|_2^2 = \ell_i^2(A).$$

□

**Lemma 5.** Suppose $A \in \mathbb{R}^{n\times\prod_{d=1}^2 p_d}$ has leverage scores $\ell_i^2(A)$ for all $i \in [n]$. Then for any $v_i^{(r)} \in \mathbb{R}^{p_2}$, $i \in [2]$, $r \in [R]$ with $R \le p_2/2$, the leverage scores of $A^{\left\{v_i^{(r)}\right\}} = \left[A^{v_1^{(1)}}, \dots, A^{v_1^{(R)}}, A^{v_2^{(1)}}, \dots, A^{v_2^{(R)}}\right] \in \mathbb{R}^{n\times 2Rp_1}$ are bounded by $\ell_i^2\left(A^{\left\{v_i^{(r)}\right\}}\right) \le \ell_i^2(A)$.

*Proof.* Let $Z$ have orthonormal columns and have the same span as the column space of $A$. Then we have $\ell_i^2(A) = \|e_i^\top Z\|_2^2$ for all $i \in [n]$. Since the column space of $A^{\left\{v_i^{(r)}\right\}}$ is a subspace of the column space of $A$, as the column space of each $A^{v_i^{(r)}}$ is a subspace of the column space of $A$, we can always find a column sub-matrix $Z_1 \in \mathbb{R}^{n\times 2Rp_1}$ of $Z$ such that $Z_1$ spans the column space of $A^{\left\{v_i^{(r)}\right\}}$. Therefore, for each $i \in [n]$, we have

$$\ell_i^2\left(A^{\left\{v_i^{(r)}\right\}}\right) = \|e_i^\top Z_1\|_2^2 \le \|e_i^\top Z\|_2^2 = \ell_i^2(A).$$

□

**Lemma 6.** For any $v_1, v_2 \in \mathcal{B}_{p_2}$, suppose $\langle v_1, v_2\rangle = 0$, and $\overline{v}_1, \overline{v}_2 \in \mathcal{B}_{p_2}$ are vectors such that $\|v_1 - \overline{v}_1\|_2 \le \eta_0$ and $\|v_2 - \overline{v}_2\|_2 \le \eta_0$. Then we have

$$\rho_{\mathrm{Fin}}([A^{v_1}, A^{v_2}], [A^{\overline{v}_1}, A^{\overline{v}_2}]) \le 2\kappa(A)\eta_0.$$

*Proof.* Denote $A^{v_1,v_2} = [A^{v_1}, A^{v_2}]$. From a perturbation bound for orthogonal projections given in Li et al. (2013a), we have

$$\rho_{\mathrm{Fin}}(A^{v_1,v_2}, A^{\overline{v}_1,\overline{v}_2}) \le \frac{\|A^{v_1,v_2} - A^{\overline{v}_1,\overline{v}_2}\|_2}{\sigma_{\min}(A^{v_1,v_2})}. \tag{42}$$

We first provide an upper bound on the numerator as

$$\begin{aligned} \|A^{v_1,v_2} - A^{\overline{v}_1,\overline{v}_2}\|_2 &= \left\|\left[\sum_{i=1}^{p_2} A^{(i)}(v_{1,i} - \overline{v}_{1,i}), \sum_{i=1}^{p_2} A^{(i)}(v_{2,i} - \overline{v}_{2,i})\right]\right\|_2 \\ &\le \left\|\sum_{i=1}^{p_2} A^{(i)}(v_{1,i} - \overline{v}_{1,i})\right\|_2 + \left\|\sum_{i=1}^{p_2} A^{(i)}(v_{2,i} - \overline{v}_{2,i})\right\|_2 \le 2\sigma_{\max}(A)\eta_0, \end{aligned} \tag{43}$$

where the last inequality is from Lemma 2.

Next, we provide a lower bound on the denominator. Let $[u_1^\top, u_2^\top]^\top$ be a unit vector corresponding to the smallest singular value of $A^{v_1,v_2}$, where $u_1, u_2 \in \mathbb{R}^{p_1}$. Then we have

$$
\begin{aligned}
\sigma_{\min}(A^{v_1,v_2}) = \left\| A^{v_1,v_2} \begin{bmatrix} u_1 \\ u_2 \end{bmatrix} \right\|_2 &= \|A(v_1 \otimes u_1 + v_2 \otimes u_2)\|_2 \geq \sigma_{\min}(A)\|v_1 \otimes u_1 + v_2 \otimes u_2\|_2 \\
&= \sigma_{\min}(A)\sqrt{\|v_1 \otimes u_1\|_2^2 + \|v_2 \otimes u_2\|_2^2 + 2\langle v_1 \otimes u_1, v_2 \otimes u_2\rangle} \\
&= \sigma_{\min}(A)\sqrt{\|u_1\|_2^2 + \|u_2\|_2^2 + 2\sum_{i=1}^{p_2}\sum_{j=1}^{p_1} v_{1,i}u_{1,j}v_{2,i}u_{2,j}} \\
&= \sigma_{\min}(A)\sqrt{1 + 2\langle v_1, v_2\rangle\langle u_1, u_2\rangle} = \sigma_{\min}(A),
\end{aligned} \tag{44}
$$

where the last equality is from the condition $\langle v_1, v_2\rangle = 0$. We finish the proof by combining (42), (43), and (44).

□

**Lemma 7.** For all $i \in [2]$ and $r \in [R]$, $v_i^{(r)} \in \mathcal{B}_{p_2}$. Suppose for all $i \in [2]$, $r \in [R]$, $q \in [R]\backslash\{r\}$, we have $\langle v_i^{(r)}, v_i^{(q)}\rangle = \langle v_1^{(r)}, v_2^{(r)}\rangle = 0$. Further suppose for all $i \in [2]$ and $r \in [R]$, $\overline{v}_i^{(r)} \in \mathcal{B}_{p_2}$ is a vector such that $\|v_i^{(r)} - \overline{v}_i^{(r)}\|_2 \leq \eta_0$. Denote $A^{\left\{v_i^{(r)}\right\}} = \left[A^{v_1^{(1)}}, \ldots, A^{v_1^{(R)}}, A^{v_2^{(1)}}, \ldots, A^{v_2^{(R)}}\right]$. Then we have

$$
\rho_{\mathrm{Fin}}\left(A^{\left\{v_i^{(r)}\right\}}, A^{\left\{\overline{v}_i^{(r)}\right\}}\right) \leq 2R\kappa(A)\eta_0.
$$

*Proof.* From the perturbation bound for orthogonal projection given in Li et al. (2013a), we have

$$
\rho_{\mathrm{Fin}}\left(A^{\left\{v_i^{(r)}\right\}}, A^{\left\{\overline{v}_i^{(r)}\right\}}\right) \leq \frac{\left\| A^{\left\{v_i^{(r)}\right\}} - A^{\left\{\overline{v}_i^{(r)}\right\}} \right\|_2}{\sigma_{\min}\left(A^{\left\{v_i^{(r)}\right\}}\right)}. \tag{45}
$$

We first upper bound the numerator as

$$
\begin{aligned}
&\left\| A^{\left\{v_i^{(r)}\right\}} - A^{\left\{\overline{v}_i^{(r)}\right\}} \right\|_2 \\
&= \left\| \left[ \sum_{j=1}^{p_2} A_j\left(v_{1,j}^{(1)} - \overline{v}_{1,j}^{(1)}\right), \ldots, \sum_{j=1}^{p_2} A_j\left(v_{1,j}^{(R)} - \overline{v}_{1,j}^{(R)}\right), \sum_{j=1}^{p_2} A_j\left(v_{2,j}^{(1)} - \overline{v}_{2,j}^{(1)}\right), \ldots, \sum_{j=1}^{p_2} A_j\left(v_{2,j}^{(R} - \overline{v}_{2,j}^{(R)}\right) \right] \right\|_2 \\
&\leq \sum_{r=1}^{R} \left\| \sum_{j=1}^{p_2} A_j\left(v_{1,j}^{(r)} - \overline{v}_{1,j}^{(r)}\right) \right\|_2 + \sum_{r=1}^{R} \left\| \sum_{j=1}^{p_2} A_j\left(v_{2,j}^{(r)} - \overline{v}_{2,j}^{(r)}\right) \right\|_2 \leq 2R\sigma_{\max}(A)\eta_0,
\end{aligned} \tag{46}
$$

where the last inequality is from Lemma 2.

Next, we provide a lower bound on the denominator. Let $\left[u_1^{(1)\top},\ldots,u_1^{(R)\top},u_2^{(1)\top},\ldots,u_2^{(R)\top}\right]^\top \in \mathbb{R}^{2Rp_1}$ be a unit vector corresponding to the smallest singular value of $A^{\left\{v_i^{(r)}\right\}}$, where $u_i^{(r)} \in \mathbb{R}^{p_1}$ for all $i \in [2]$ and $r \in [R]$. Then we have

$$
\begin{aligned}
&\sigma_{\min}\left(A^{\left\{v_i^{(r)}\right\}}\right) = \left\|A^{\left\{v_i^{(r)}\right\}}\left[u_1^{(1)\top},\ldots,u_1^{(R)\top},u_2^{(1)\top},\ldots,u_2^{(R)\top}\right]^\top\right\|_2 = \left\|A\cdot\left(\sum_{r=1}^{R} v_1^{(r)}\otimes u_1^{(r)} + v_2^{(r)}\otimes u_2^{(r)}\right)\right\|_2 \\
&\geq \sigma_{\min}(A)\left\|\sum_{r=1}^{R}\left(v_1^{(r)}\otimes u_1^{(r)} + v_2^{(r)}\otimes u_2^{(r)}\right)\right\|_2 \\
&= \sigma_{\min}(A)\sqrt{\sum_{r=1}^{R}\left(\left\|u_1^{(r)}\right\|_2^2+\left\|u_2^{(r)}\right\|_2^2\right)+2\sum_{r=1}^{R}\sum_{j=1}^{p_2}\sum_{k=1}^{p_1} v_{1,j}^{(r)}u_{1,k}^{(r)}v_{2,j}^{(r)}u_{2,k}^{(r)}+2\sum_{i=1}^{2}\sum_{r=1}^{R-1}\sum_{q=r+1}^{R}\sum_{j=1}^{p_2}\sum_{k=1}^{p_1} v_{i,j}^{(r)}u_{i,k}^{(r)}v_{i,j}^{(q)}u_{i,k}^{(q)}} \\
&= \sigma_{\min}(A)\sqrt{1+2\sum_{r=1}^{R}\langle v_1^{(r)},v_2^{(r)}\rangle\langle u_1^{(r)},u_2^{(r)}\rangle+2\sum_{i=1}^{2}\sum_{r=1}^{R-1}\sum_{q=r+1}^{R}\langle v_i^{(r)},v_i^{(q)}\rangle\langle u_i^{(r)},u_i^{(q)}\rangle} = \sigma_{\min}(A),
\end{aligned} \tag{47}
$$

where the last equality uses the conditions that for all $i \in [2]$ and $r \in [R]$, $\langle v_i^{(r)}, v_i^{(q)}\rangle = \langle v_1^{(r)}, v_2^{(r)}\rangle = 0$ for $q \in [R]\backslash\{r\}$. We finish the proof by combining (45), (46), and (47).

□

**Lemma 8.** For all $d \in [D]\backslash\{1\}$, $\theta_d, \phi_d \in \mathcal{B}_{p_d}$. Suppose there exists an $i \in [D]\backslash\{1\}$ such that $\langle\theta_i,\phi_i\rangle = 0$. Further suppose for all $d \in [D]\backslash\{1\}$, $\overline{\theta}_d, \overline{\phi}_d \in \mathcal{B}_{p_d}$ are vectors such that $\|\theta_d - \overline{\theta}_d\|_2 \leq \eta_0$ and $\|\phi_d - \overline{\phi}_d\|_2 \leq \eta_0$. Then we have

$$\rho_{\text{Fin}}\left(\left[A^{\{\theta_{\backslash 1}\}}, A^{\{\phi_{\backslash 1}\}}\right], \left[A^{\{\overline{\theta}_{\backslash 1}\}}, A^{\{\overline{\phi}_{\backslash 1}\}}\right]\right) \leq 2\kappa(A)\left((1+\eta_0)^{D-1}-1\right).$$

*Proof.* Let $A^{\vartheta_{\backslash 1},\varphi_{\backslash 1}} = \left[A^{\{\theta_{\backslash 1}\}}, A^{\{\phi_{\backslash 1}\}}\right] \in \mathbb{R}^{n\times 2p_1}$. From the perturbation bound for orthogonal projection given in Li et al. (2013a), we have

$$\rho_{\text{Fin}}\left(A^{\vartheta_{\backslash 1},\varphi_{\backslash 1}}, A^{\overline{\vartheta},\overline{\varphi}}\right) \leq \frac{\left\|A^{\vartheta_{\backslash 1},\varphi_{\backslash 1}} - A^{\overline{\vartheta},\overline{\varphi}}\right\|_2}{\sigma_{\min}(A^{\vartheta_{\backslash 1},\varphi_{\backslash 1}})}. \tag{48}$$

We denote $\sum_{j_2\cdots j_D} = \sum_{j_D=1}^{p_D}\cdots\sum_{j_2=1}^{p_2}$. We first provide an upper bound on the numerator:

$$
\begin{aligned}
&\left\|A^{\vartheta_{\backslash 1},\varphi_{\backslash 1}} - A^{\overline{\vartheta},\overline{\varphi}}\right\|_2 \\
&= \left\|\left[\sum_{j_2\cdots j_D} A^{(j_D,\ldots,j_2)}\cdot\left(\theta_{D,j_D}\cdots\theta_{2,j_2} - \overline{\theta}_{D,j_D}\cdots\overline{\theta}_{2,j_2}\right), \sum_{j_2\cdots j_D} A^{(j_D,\ldots,j_2)}\cdot\left(\phi_{D,j_D}\cdots\phi_{2,j_2} - \overline{\phi}_{D,j_D}\cdots\overline{\phi}_{2,j_2}\right)\right]\right\|_2 \\
&\leq \left\|\sum_{j_2\cdots j_D} A^{(j_D,\ldots,j_2)}\cdot\left(\theta_{D,j_D}\cdots\theta_{2,j_2} - \overline{\theta}_{D,j_D}\cdots\overline{\theta}_{2,j_2}\right)\right\|_2 + \left\|\sum_{j_2\cdots j_D} A^{(j_D,\ldots,j_2)}\cdot\left(\phi_{D,j_D}\cdots\phi_{2,j_2} - \overline{\phi}_{D,j_D}\cdots\overline{\phi}_{2,j_2}\right)\right\|_2 \\
&\leq \sigma_{\max}(A)\cdot\left(\left\|\theta_D\otimes\cdots\otimes\theta_2 - \overline{\theta}_D\otimes\cdots\otimes\overline{\theta}_2\right\|_2 + \left\|\phi_D\otimes\cdots\otimes\phi_2 - \overline{\phi}_D\otimes\cdots\otimes\overline{\phi}_2\right\|_2\right) \\
&\leq 2\sigma_{\max}(A)\left((1+\eta_0)^{D-1}-1\right),
\end{aligned} \tag{49}
$$

where the second inequality is from Lemma 2 and the last inequality is from Lemma 3.

Next, we provide a lower bound on the denominator. Let $[u_1^\top, u_2^\top]^\top$ be a unit vector corresponding to the smallest singular value of $A^{\vartheta_{\backslash 1}, \varphi_{\backslash 1}}$, where $u_1, u_2 \in \mathbb{R}^{p_1}$. Then we have

$$\begin{aligned}
&\sigma_{\min}\left(A^{\vartheta_{\backslash 1}, \varphi_{\backslash 1}}\right) = \left\| A^{\vartheta_{\backslash 1}, \varphi_{\backslash 1}} \begin{bmatrix} u_1 \\ u_2 \end{bmatrix} \right\|_2 = \|A(\theta_D \otimes \cdots \otimes \theta_2 \otimes u_1 + \phi_D \otimes \cdots \otimes \phi_2 \otimes u_2)\|_2 \\
&\geq \sigma_{\min}(A) \|\theta_D \otimes \cdots \otimes \theta_2 \otimes u_1 + \phi_D \otimes \cdots \otimes \phi_2 \otimes u_2\|_2 \\
&= \sigma_{\min}(A) \sqrt{\|\theta_D \otimes \cdots \otimes \theta_2 \otimes u_1\|_2^2 + \|\phi_D \otimes \cdots \otimes \phi_2 \otimes u_2\|_2^2 + 2\langle \theta_D \otimes \cdots \otimes \theta_2 \otimes u_1, \phi_D \otimes \cdots \otimes \phi_2 \otimes u_2 \rangle} \\
&= \sigma_{\min}(A) \sqrt{\|u_1\|_2^2 + \|u_2\|_2^2 + 2 \sum_{j_2 \cdots j_D} \sum_{j_1=1}^{p_1} \theta_{D,j_D} \cdots \theta_{2,j_2} u_{1,j_1} \cdot \phi_{D,j_D} \cdots \phi_{2,j_2} u_{2,j_1}} \\
&= \sigma_{\min}(A) \sqrt{1 + 2\langle \theta_D, \phi_D \rangle \cdots \langle \theta_2, \phi_2 \rangle \langle u_1, u_2 \rangle} = \sigma_{\min}(A), \qquad (50)
\end{aligned}$$

where the last inequality is from $\langle \theta_i, \phi_i \rangle = 0$ for some $i \in \{2, \ldots, D\}$. We finish the proof by combining (48), (49) and (50).

□

**Lemma 9.** For all $d \in [D] \backslash \{1\}$ and $r \in [R]$, $\theta_d^{(r)}, \phi_d^{(r)} \in \mathcal{B}_{p_d}$. Suppose that for any $r, q \in [R]$, there exists an $i \in [D] \backslash \{1\}$ such that $\langle \theta_i^{(r)}, \phi_i^{(q)} \rangle = 0$, and further, for all $r \in [R-1]$, $q \in [R] \backslash [r]$, there exist $j, k \in [D] \backslash \{1\}$ such that $\langle \theta_j^{(r)}, \theta_j^{(q)} \rangle = 0$ and $\langle \phi_k^{(r)}, \phi_k^{(q)} \rangle = 0$. Further suppose for all $d \in [D] \backslash \{1\}$ and $r \in [R]$, $\overline{\theta}_d^{(r)}, \overline{\phi}_d^{(r)} \in \mathcal{B}_{p_d}$ are vectors such that $\|\theta_d^{(r)} - \overline{\theta}_d^{(r)}\|_2 \leq \eta_0$ and $\|\phi_d^{(r)} - \overline{\phi}_d^{(r)}\|_2 \leq \eta_0$. Then we have

$$\rho_{\text{Fin}}\left(\left[A^{\left\{\theta_{\backslash 1}^{(r)}\right\}}, A^{\left\{\phi_{\backslash 1}^{(r)}\right\}}\right], \left[A^{\left\{\overline{\theta}_{\backslash 1}^{(r)}\right\}}, A^{\left\{\overline{\phi}_{\backslash 1}^{(r)}\right\}}\right]\right) \leq 2R\kappa(A)\left((1+\eta_0)^{D-1} - 1\right).$$

*Proof.* Denote $A^{\left\{\vartheta_{\backslash 1}^{(r)}, \varphi_{\backslash 1}^{(r)}\right\}} = \left[A^{\left\{\theta_{\backslash 1}^{(r)}\right\}}, A^{\left\{\phi_{\backslash 1}^{(r)}\right\}}\right] \in \mathbb{R}^{n \times 2Rp_1}$. From the perturbation bound on orthogonal projection given in Li et al. (2013a), we have

$$\rho_{\text{Fin}}\left(A^{\left\{\vartheta_{\backslash 1}^{(r)}, \varphi_{\backslash 1}^{(r)}\right\}}, A^{\left\{\overline{\vartheta}_{\backslash 1}^{(r)}, \overline{\varphi}_{\backslash 1}^{(r)}\right\}}\right) \leq \frac{\left\| A^{\left\{\vartheta_{\backslash 1}^{(r)}, \varphi_{\backslash 1}^{(r)}\right\}} - A^{\left\{\overline{\vartheta}_{\backslash 1}^{(r)}, \overline{\varphi}_{\backslash 1}^{(r)}\right\}} \right\|_2}{\sigma_{\min}\left(A^{\left\{\vartheta_{\backslash 1}^{(r)}, \varphi_{\backslash 1}^{(r)}\right\}}\right)}. \qquad (51)$$

We denote $\sum_{j_2\cdots j_D} = \sum_{j_D=1}^{p_D}\cdots\sum_{j_2=1}^{p_2}$. We first upper bound the numerator as

$$
\begin{aligned}
&\left\| A^{\left\{\vartheta_{\setminus 1}^{(r)},\varphi_{\setminus 1}^{(r)}\right\}} - A^{\left\{\overline{\vartheta}_{\setminus 1}^{(r)},\overline{\varphi}_{\setminus 1}^{(r)}\right\}} \right\|_2 \\
&= \Bigg\| \Bigg[ \sum_{j_2\cdots j_D} A^{(j_D,\dots,j_2)}\cdot\left(\theta_{D,j_D}^{(1)}\cdots\theta_{2,j_2}^{(1)} - \overline{\theta}_{D,j_D}^{(1)}\cdots\overline{\theta}_{2,j_2}^{(1)}\right),\dots,\sum_{j_2\cdots j_D} A^{(j_D,\dots,j_2)}\cdot\left(\theta_{D,j_D}^{(R)}\cdots\theta_{2,j_2}^{(R)} - \overline{\theta}_{D,j_D}^{(R)}\cdots\overline{\theta}_{2,j_2}^{(R)}\right), \\
&\qquad \sum_{j_2\cdots j_D} A^{(j_D,\dots,j_2)}\cdot\left(\phi_{D,j_D}^{(1)}\cdots\phi_{2,j_2}^{(1)} - \overline{\phi}_{D,j_D}^{(1)}\cdots\overline{\phi}_{2,j_2}^{(1)}\right),\dots,\sum_{j_2\cdots j_D} A^{(j_D,\dots,j_2)}\cdot\left(\phi_{D,j_D}^{(R)}\cdots\phi_{2,j_2}^{(R)} - \overline{\phi}_{D,j_D}^{(R)}\cdots\overline{\phi}_{2,j_2}^{(R)}\right)\Bigg]\Bigg\|_2 \\
&\le \sum_{r=1}^{R} \left\| \sum_{j_2\cdots j_D} A^{(j_D,\dots,j_2)}\cdot\left(\theta_{D,j_D}^{(r)}\cdots\theta_{2,j_2}^{(r)} - \overline{\theta}_{D,j_D}^{(r)}\cdots\overline{\theta}_{2,j_2}^{(r)}\right)\right\|_2 + \left\| \sum_{j_2\cdots j_D} A^{(j_D,\dots,j_2)}\cdot\left(\phi_{D,j_D}^{(r)}\cdots\phi_{2,j_2}^{(r)} - \overline{\phi}_{D,j_D}^{(r)}\cdots\overline{\phi}_{2,j_2}^{(r)}\right)\right\|_2 \\
&\le \sigma_{\max}(A)\cdot\left(\sum_{r=1}^{R}\left\|\theta_D^{(r)}\otimes\cdots\otimes\theta_2^{(r)} - \overline{\theta}_D^{(r)}\otimes\cdots\otimes\overline{\theta}_2^{(r)}\right\|_2 + \left\|\phi_D^{(r)}\otimes\cdots\otimes\phi_2^{(r)} - \overline{\phi}_D^{(r)}\otimes\cdots\otimes\overline{\phi}_2^{(r)}\right\|_2\right) \\
&\le 2R\sigma_{\max}(A)\left((1+\eta_0)^{D-1}-1\right), \qquad (52)
\end{aligned}
$$

where the second inequality is from Lemma 2 and the last inequality is from Lemma 3.

Next, we lower bound the denominator. Let $\left[u_1^{(1)\top},\dots,u_1^{(R)\top},u_2^{(1)\top},\dots,u_2^{(R)\top}\right]^\top \in \mathbb{R}^{2Rp_1}$ be a unit vector corresponding to the smallest singular value of $A^{\left\{\vartheta_{\setminus 1}^{(r)},\varphi_{\setminus 1}^{(r)}\right\}}$, where $u_i^{(r)} \in \mathbb{R}^{p_1}$ for all $i \in [2]$ and

$r \in [R]$. Then we have

$$\sigma_{\min}\left(A^{\left\{\vartheta_{\backslash 1}^{(r)},\varphi_{\backslash 1}^{(r)}\right\}}\right) = \left\|A^{\left\{\vartheta_{\backslash 1}^{(r)},\varphi_{\backslash 1}^{(r)}\right\}}\left[u_1^{(1)\top},\ldots,u_1^{(R)\top},u_2^{(1)\top},\ldots,u_2^{(R)\top}\right]^\top\right\|_2$$
$$= \left\|A\cdot\left(\sum_{r=1}^R \theta_D^{(r)}\otimes\cdots\otimes\theta_2^{(r)}\otimes u_1^{(r)} + \phi_D^{(r)}\otimes\cdots\otimes\phi_2^{(r)}\otimes u_2^{(r)}\right)\right\|_2$$
$$\geq \sigma_{\min}(A)\left\|\sum_{r=1}^R \theta_D^{(r)}\otimes\cdots\otimes\theta_2^{(r)}\otimes u_1^{(r)} + \phi_D^{(r)}\otimes\cdots\otimes\phi_2^{(r)}\otimes u_2^{(r)}\right\|_2$$
$$= \sigma_{\min}(A)\sqrt{\sum_{r=1}^R\left(\left\|u_1^{(r)}\right\|_2^2+\left\|u_2^{(r)}\right\|_2^2\right)+2\sum_{r=1}^R\sum_{q=1}^R\sum_{j_1\cdots j_D}\theta_{D,j_D}^{(r)}\cdots\theta_{2,j_2}^{(r)}u_{1,j_1}^{(r)}\cdot\phi_{D,j_D}^{(q)}\cdots\phi_{2,j_2}^{(q)}u_{2,j_1}^{(q)}}$$
$$\overline{+2\sum_{r=1}^{R-1}\sum_{q=r+1}^R\sum_{j_1\cdots j_D}\left(\theta_{D,j_D}^{(r)}\cdots\theta_{2,j_2}^{(r)}u_{1,j_1}^{(r)}\cdot\theta_{D,j_D}^{(q)}\cdots\theta_{2,j_2}^{(q)}u_{1,j_1}^{(q)}+\phi_{D,j_D}^{(r)}\cdots\phi_{2,j_2}^{(r)}u_{2,j_1}^{(r)}\cdot\phi_{D,j_D}^{(q)}\cdots\phi_{2,j_2}^{(q)}u_{2,j_1}^{(q)}\right)}$$
$$= \sigma_{\min}(A)\sqrt{1+2\sum_{r=1}^R\sum_{q=1}^R\langle\theta_D^{(r)},\phi_D^{(q)}\rangle\cdots\langle\theta_2^{(r)},\phi_2^{(q)}\rangle\langle u_1^{(r)},u_2^{(q)}\rangle}$$
$$\overline{+2\sum_{r=1}^{R-1}\sum_{q=r+1}^R\left(\langle\theta_D^{(r)},\theta_D^{(q)}\rangle\cdots\langle\theta_2^{(r)},\theta_2^{(q)}\rangle\langle u_1^{(r)},u_1^{(q)}\rangle+\langle\phi_D^{(r)},\phi_D^{(q)}\rangle\cdots\langle\phi_2^{(r)},\phi_2^{(q)}\rangle\langle u_2^{(r)},u_2^{(q)}\rangle\right)}$$
$$= \sigma_{\min}(A), \tag{53}$$

where the last inequality is from the conditions on $\theta_d^{(r)}$ and $\phi_d^{(r)}$. We finish the proof by combining (51), (52), and (53). □

**Lemma 10.** For all $d \in [D]\backslash\{1\}$ and $r \in [R]$, $\theta_d^{(r_d)}, \phi_d^{(r_d)} \in \mathcal{B}_{p_d}$. Suppose that for any $r_d, q_d \in [R_d]$, $d \in [R]\backslash\{1\}$, there exists an $i \in [D]\backslash\{1\}$ such that $\langle\theta_i^{(r)}, \phi_i^{(q)}\rangle = 0$, and for all $r \in [R-1]$, $q \in [R]\backslash[r]$, there exist $j,k \in [D]\backslash\{1\}$ such that $\langle\theta_j^{(r)}, \theta_j^{(q)}\rangle = 0$ and $\langle\phi_k^{(r)}, \phi_k^{(q)}\rangle = 0$. Further suppose for all $d \in [D]\backslash\{1\}$ and $r \in [R]$, $\overline{\theta}_d^{(r)}, \overline{\phi}_d^{(r)} \in \mathcal{B}_{p_d}$ are vectors such that $\|\theta_d^{(r)} - \overline{\theta}_d^{(r)}\|_2 \leq \eta_0$ and $\|\phi_d^{(r)} - \overline{\phi}_d^{(r)}\|_2 \leq \eta_0$. Then we have

$$\rho_{\mathrm{Fin}}\left(\left[A^{\left\{\theta_{\backslash 1}^{(r)}\right\}}, A^{\left\{\phi_{\backslash 1}^{(r)}\right\}}\right], \left[A^{\left\{\overline{\theta}_{\backslash 1}^{(r)}\right\}}, A^{\left\{\overline{\phi}_{\backslash 1}^{(r)}\right\}}\right]\right) \leq 2\kappa(A)\left((1+\eta_0)^{D-1}-1\right)\sqrt{\prod_{d=2}^D R_d}.$$

*Proof.* Denote $A^{\left\{\vartheta_{\backslash 1}^{\{r_d\}},\varphi_{\backslash 1}^{\{r_d\}}\right\}} = \left[A^{\left\{\theta_{\backslash 1}^{\{r_d\}}\right\}}, A^{\left\{\phi_{\backslash 1}^{\{r_d\}}\right\}}\right] \in \mathbb{R}^{n\times 2R_1p_1}$. From the perturbation bound for or-thogonal projection given in Li et al. (2013a), we have

$$\rho_{\text{Fin}}\left(A^{\left\{\vartheta_{\backslash 1}^{\{r_d\}},\varphi_{\backslash 1}^{\{r_d\}}\right\}},A^{\left\{\overline{\vartheta}_{\backslash 1}^{\{r_d\}},\overline{\varphi}_{\backslash 1}^{\{r_d\}}\right\}}\right)\leq\frac{\left\|A^{\left\{\vartheta_{\backslash 1}^{\{r_d\}},\varphi_{\backslash 1}^{\{r_d\}}\right\}}-A^{\left\{\overline{\vartheta}_{\backslash 1}^{\{r_d\}},\overline{\varphi}_{\backslash 1}^{\{r_d\}}\right\}}\right\|_2}{\sigma_{\min}\left(A^{\left\{\vartheta_{\backslash 1}^{\{r_d\}},\varphi_{\backslash 1}^{\{r_d\}}\right\}}\right)}. \tag{54}$$

We denote $\sum_{j_2\cdots j_D}=\sum_{j_D=1}^{p_D}\cdots\sum_{j_2=1}^{p_2}$. We first upper bound the numerator as

$$\begin{aligned}
&\left\|A^{\left\{\vartheta_{\backslash 1}^{\{r_d\}},\varphi_{\backslash 1}^{\{r_d\}}\right\}}-A^{\left\{\overline{\vartheta}_{\backslash 1}^{\{r_d\}},\overline{\varphi}_{\backslash 1}^{\{r_d\}}\right\}}\right\|_2\\
&=\Bigg\|\Bigg[\sum_{j_2\cdots j_D}A^{(j_D,\dots,j_2)}\cdot\left(\theta_{D,j_D}^{(1)}\cdots\theta_{2,j_2}^{(1)}-\overline{\theta}_{D,j_D}^{(1)}\cdots\overline{\theta}_{2,j_2}^{(1)}\right),\dots,\sum_{j_2\cdots j_D}A^{(j_D,\dots,j_2)}\cdot\left(\theta_{D,j_D}^{(1)}\cdots\theta_{2,j_2}^{(R_2)}-\overline{\theta}_{D,j_D}^{(1)}\cdots\overline{\theta}_{2,j_2}^{(R_2)}\right),\dots,\\
&\qquad\sum_{j_2\cdots j_D}A^{(j_D,\dots,j_2)}\cdot\left(\phi_{D,j_D}^{(R_D)}\cdots\phi_{2,j_2}^{(1)}-\overline{\phi}_{D,j_D}^{(R_D)}\cdots\overline{\phi}_{2,j_2}^{(1)}\right),\dots,\sum_{j_2\cdots j_D}A^{(j_D,\dots,j_2)}\cdot\left(\phi_{D,j_D}^{(R_D)}\cdots\phi_{2,j_2}^{(R_2)}-\overline{\phi}_{D,j_D}^{(R_D)}\cdots\overline{\phi}_{2,j_2}^{(R_2)}\right)\Bigg]\Bigg\|_2\\
&\leq\sum_{r_2=1}^{R_2}\cdots\sum_{r_D=1}^{R_D}\left\|\sum_{j_2\cdots j_D}A^{(j_D,\dots,j_2)}\cdot\left(\theta_{D,j_D}^{(r_D)}\cdots\theta_{2,j_2}^{(r_2)}-\overline{\theta}_{D,j_D}^{(r_D)}\cdots\overline{\theta}_{2,j_2}^{(r_2)}\right)\right\|_2\\
&\qquad\qquad+\left\|\sum_{j_2\cdots j_D}A^{(j_D,\dots,j_2)}\cdot\left(\phi_{D,j_D}^{(r_D)}\cdots\phi_{2,j_2}^{(r_2)}-\overline{\phi}_{D,j_D}^{(r_D)}\cdots\overline{\phi}_{2,j_2}^{(r_2)}\right)\right\|_2\\
&\leq\sigma_{\max}(A)\cdot\Bigg(\sum_{r_2=1}^{R_2}\cdots\sum_{r_D=1}^{R_D}\left\|\theta_D^{(r_D)}\otimes\cdots\otimes\theta_2^{(r_2)}-\overline{\theta}_D^{(r_D)}\otimes\cdots\otimes\overline{\theta}_2^{(r_2)}\right\|_2\\
&\qquad\qquad+\left\|\phi_D^{(r_D)}\otimes\cdots\otimes\phi_2^{(r_2)}-\overline{\phi}_D^{(r_D)}\otimes\cdots\otimes\overline{\phi}_2^{(r_2)}\right\|_2\Bigg)\\
&\leq 2\prod_{d=2}^{D}R_d\cdot\sigma_{\max}(A)\left((1+\eta_0)^{D-1}-1\right),
\end{aligned}\tag{55}$$

where the second inequality is from Lemma 2 and the last inequality is from Lemma 3.

Next, we provide a lower bound on the denominator. Let $\left[u_1^{(1)\top},\dots,u_1^{(R_1)\top},u_2^{(1)\top},\dots,u_2^{(R_1)\top}\right]^\top\in\mathbb{R}^{2R_1p_1}$ be a unit vector corresponding to the smallest singular value of $A^{\left\{\vartheta_{\backslash 1}^{\{r_d\}},\varphi_{\backslash 1}^{\{r_d\}}\right\}}$, where $u_i^{(r_1)}\in$ $\mathbb{R}^{p_1}$ for all $i \in [2]$ and $r_1 \in [R_1]$. Denote $\sum_{r_1,\dots,r_D} = \sum_{r_1=1}^{R_1} \cdots \sum_{r_D=1}^{R_D}$. Then we have

$$
\begin{aligned}
&\sigma_{\min}\left(A^{\left\{\vartheta_{\backslash 1}^{\{r_d\}},\varphi_{\backslash 1}^{\{r_d\}}\right\}}\right) = \left\| A^{\left\{\vartheta_{\backslash 1}^{\{r_d\}},\varphi_{\backslash 1}^{\{r_d\}}\right\}} \left[ u_1^{(1)\top}, \dots, u_1^{(R_1)\top}, u_2^{(1)\top}, \dots, u_2^{(R_1)\top} \right]^\top \right\|_2 \\
&= \left\| A \cdot \left( \sum_{r_1,\dots,r_D} \theta_D^{(r_D)} \otimes \cdots \otimes \theta_2^{(r_2)} \otimes u_1^{(r_1)} + \phi_D^{(r_D)} \otimes \cdots \otimes \phi_2^{(r_2)} \otimes u_2^{(r_1)} \right) \right\|_2 \\
&\geq \sigma_{\min}(A) \left\| \sum_{r_1,\dots,r_D} \theta_D^{(r_D)} \otimes \cdots \otimes \theta_2^{(r_2)} \otimes u_1^{(r_1)} + \phi_D^{(r_D)} \otimes \cdots \otimes \phi_2^{(r_2)} \otimes u_2^{(r_1)} \right\|_2 \\
&= \sigma_{\min}(A) \left( \sum_{r_1,\dots,r_D} \left( \left\| u_1^{(r_1)} \right\|_2^2 + \left\| u_2^{(r_1)} \right\|_2^2 \right) + 2 \sum_{r_1,\dots,r_D} \sum_{q_1,\dots,q_D} \sum_{j_1 \cdots j_D} \theta_{D,j_D}^{(r_D)} \cdots \theta_{2,j_2}^{(r_2)} u_{1,j_1}^{(r_1)} \cdot \phi_{D,j_D}^{(q_D)} \cdots \phi_{2,j_2}^{(q_2)} u_{2,j_1}^{(q_1)} \right. \\
&\quad \left. + \sum_{r_1,\dots,r_D} \sum_{q_1,\dots,q_D} \sum_{j_1 \cdots j_D} \left( \theta_{D,j_D}^{(r_D)} \cdots \theta_{2,j_2}^{(r_2)} u_{1,j_1}^{(r_1)} \cdot \theta_{D,j_D}^{(q_D)} \cdots \theta_{2,j_2}^{(q_2)} u_{1,j_1}^{(q_1)} + \phi_{D,j_D}^{(r_D)} \cdots \phi_{2,j_2}^{(r_2)} u_{2,j_1}^{(r_1)} \cdot \phi_{D,j_D}^{(q_D)} \cdots \phi_{2,j_2}^{(q_2)} u_{2,j_1}^{(q_1)} \right) \right)^{1/2} \\
&= \sigma_{\min}(A) \left( \prod_{d=2}^D R_d + 2 \sum_{r_1,\dots,r_D} \sum_{q_1,\dots,q_D} \langle \theta_D^{(r_D)}, \phi_D^{(q_D)} \rangle \cdots \langle \theta_2^{(r_2)}, \phi_2^{(q_2)} \rangle \langle u_1^{(r_1)}, u_2^{(q_1)} \rangle \right. \\
&\quad \left. + \sum_{r_1,\dots,r_D} \sum_{q_1,\dots,q_D} \left( \langle \theta_D^{(r_D)}, \theta_D^{(q_D)} \rangle \cdots \langle \theta_2^{(r_2)}, \theta_2^{(q_2)} \rangle \langle u_1^{(r_1)}, u_1^{(q_1)} \rangle + \langle \phi_D^{(r_D)}, \phi_D^{(q_D)} \rangle \cdots \langle \phi_2^{(r_2)}, \phi_2^{(q_2)} \rangle \langle u_1^{(r_1)}, u_2^{(q_1)} \rangle \right) \right)^{1/2} \\
&= \sigma_{\min}(A) \sqrt{\prod_{d=2}^D R_d},
\end{aligned}
\tag{56}
$$

where the last inequality is from the conditions on $\theta_d^{(r)}$ and $\phi_d^{(r)}$. We finish the proof by combining (54), (55), and (56).

□